\documentclass[]{xiaomiev}

\usepackage[toc,page,header]{appendix}
\usepackage{minitoc}
\usepackage{solarized-light}
\usepackage{booktabs}
\usepackage{multirow}
\usepackage{graphicx}
\usepackage{array}
\usepackage{siunitx,array}
\usepackage[table]{xcolor}
\usepackage{makecell,array,booktabs}
\usepackage{makecell}
\usepackage{framed}

\usepackage{bbding}
\usepackage{hyperref}
\usepackage{amsmath} 
\usepackage{placeins}
\usepackage[colorinlistoftodos]{todonotes}
\usepackage{longtable}
\usepackage{hhline}
\usepackage{fancyvrb}
\usepackage{fvextra}
\usepackage{CJKutf8}
\usepackage{multicol}
\usepackage{cleveref}
\usepackage{tablefootnote}
\usepackage{threeparttable}
\usepackage{tabularx}
\usepackage{mdframed}
\usepackage{subcaption}
\usepackage[usestackEOL]{stackengine}
\usepackage[numbers]{natbib}
\newcommand{\commentout}[1]{}
\renewcommand{\paragraph}[1]{\noindent\textbf{#1.}}

\usepackage{enumitem}
\usepackage{hyperref}

\setlist[itemize]{leftmargin=15pt}
\makeatletter
\providecommand{\@bottomtitlebar}{}
\makeatother

\RequirePackage{xspace}
\makeatletter
\DeclareRobustCommand\onedot{\futurelet\@let@token\@onedot}
\def\@onedot{\ifx\@let@token.\else.\null\fi\xspace}

\makeatother
\newcommand{\Ours}{\texttt{Xiaomi-Robotics-U0}}

\title{\mbox{Xiaomi-Robotics-U0: Unified Embodied Synthesis}\\ with World Foundation Model
}
\author[\textcolor{xiaomievblue}{1}]{Xiaomi Robotics}

\abstract{
Recent foundation image and video generation models offer strong
generalization and controllability, but their direct application to
embodied scenarios is limited by requirements for multi-view consistency,
geometric coherence, and robot embodiment constraints. Existing methods
typically adapt foundation models with limited robot data, often sacrificing
visual knowledge acquired during large-scale pre-training.

We present \texttt{\Ours}, a 38-billion-parameter multimodal autoregressive
model for unified embodied synthesis. It treats embodied generation as an
extension of foundation image and video generation and jointly optimizes
text-to-image generation, image editing, embodied scene generation, embodied
transfer, and embodied video generation. This unified framework preserves the
generalization of the pre-trained world foundation model while adapting it to
embodied settings.

\texttt{\Ours} is the first model to support high-quality multi-view scene
generation across multiple robot embodiments and to introduce structured,
controllable embodied transfer for fine-grained editing while preserving
multi-view consistency and interaction dynamics. It achieves state-of-the-art
results on single-step and sequential generation tasks, outperforming
GPT-Image-2.0 in human evaluations of embodied scene generation and transfer,
ranking first on World Arena for embodied video generation, and improving the
out-of-distribution success rate of $\pi_{0.5}$ from 36.9\% to 63.2\% on
challenging real-world manipulation tasks. These results show that foundation
world models can serve both as embodied world models and scalable data engines
for embodied intelligence. Code and checkpoints are available at
\url{https://robotics.xiaomi.com/xiaomi-robotics-u0.html}.
}

\begin{document}
\maketitle
\footnotetext[1]{See Contributions section for full author list. Please send correspondence to \href{mailto:mi-robotics@xiaomi.com}{\mbox{mi-robotics@xiaomi.com}}.}

\section{Introduction}
\begin{figure*}[htbp]
    \centering
    \includegraphics[width=\textwidth]{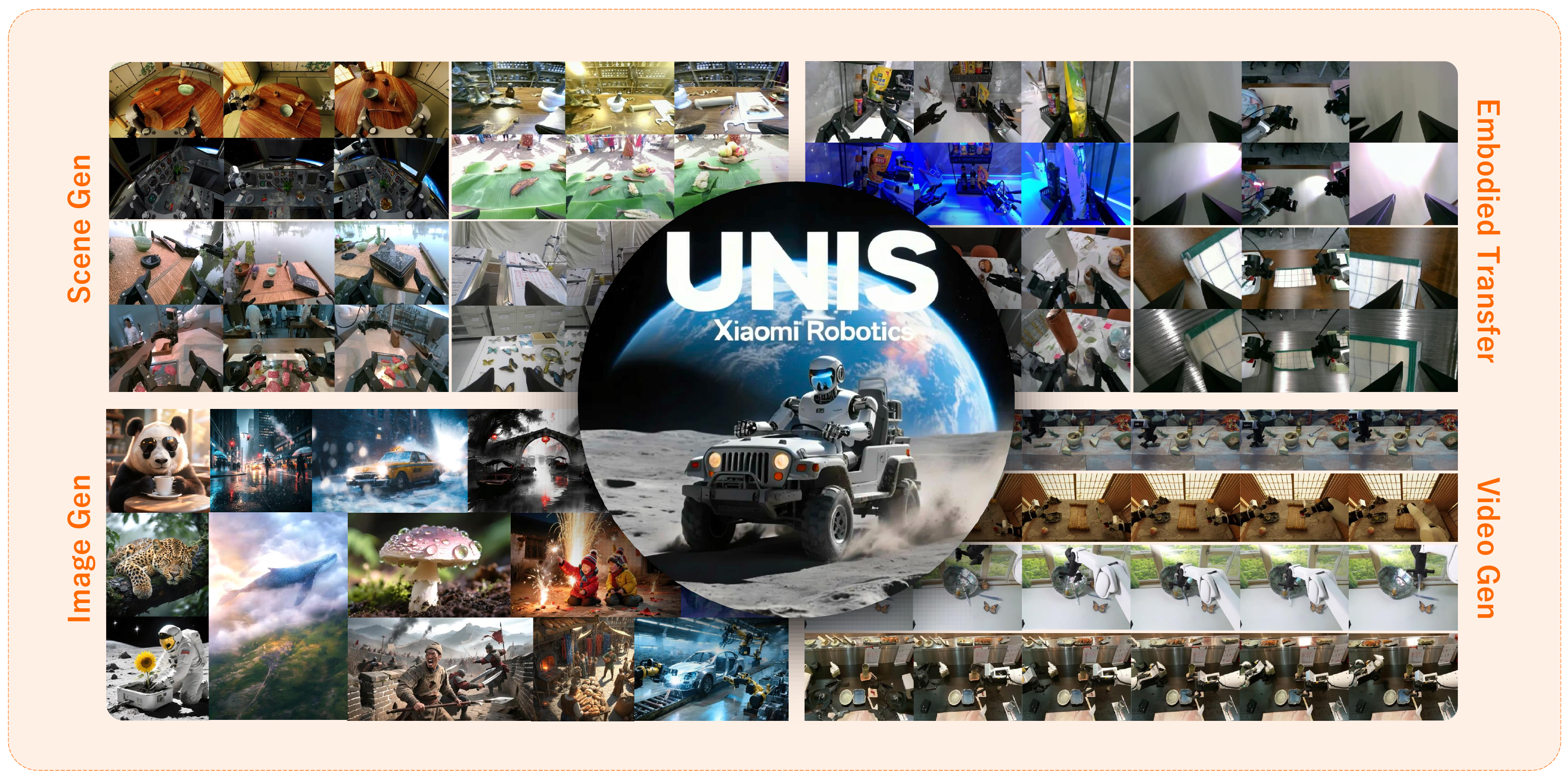}
    \caption{Embodied and general capabilities of \Ours. The rectangle corresponds to the initial observations for the same embodiment, the pairwise transfer sample, and the keyframes within a video for the embodied capabilities. All frames are referenced and generated images.}
    \label{fig:illustarte}
\end{figure*}

\begin{figure*}[htbp]
    \centering
    \includegraphics[width=\textwidth]{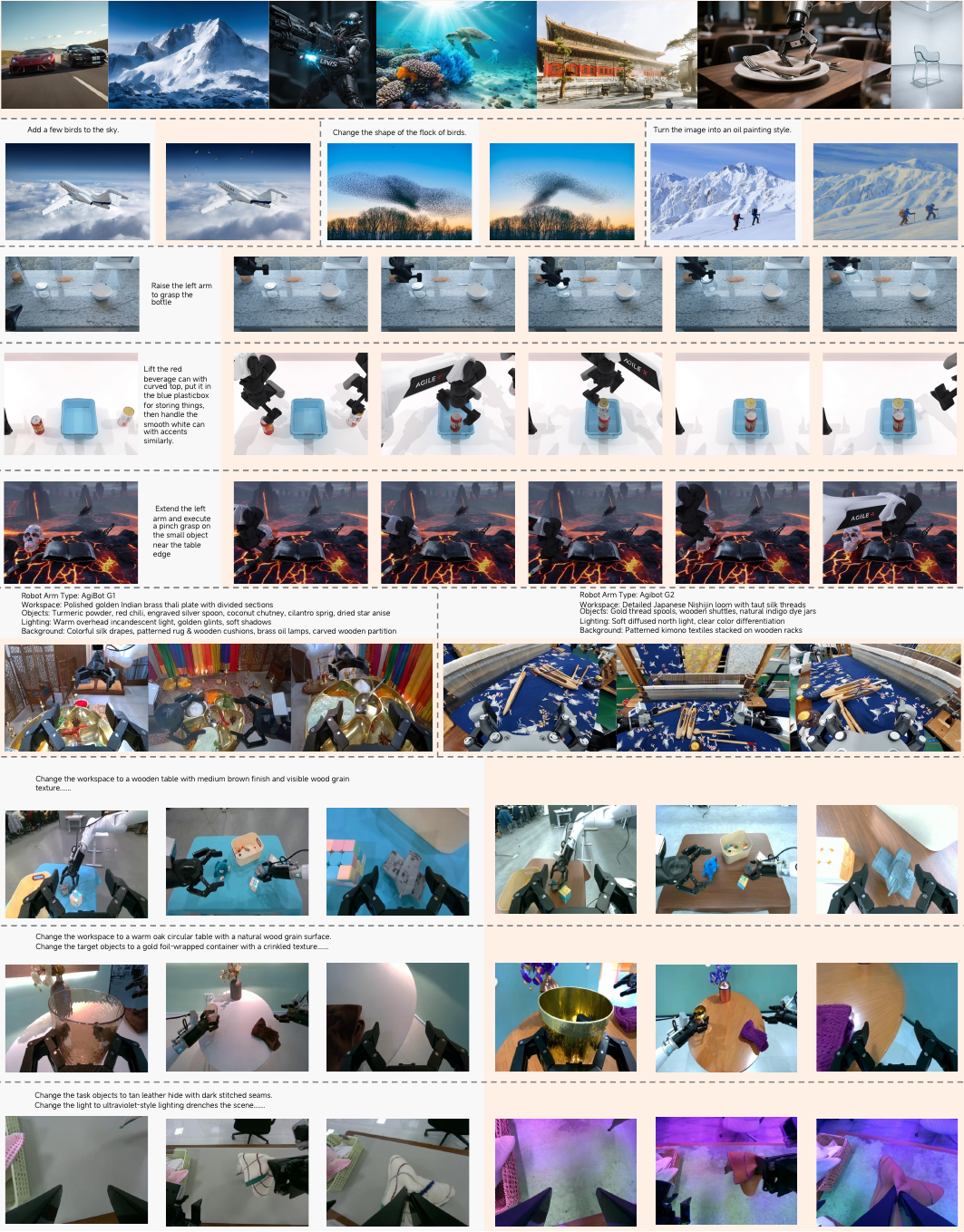}
    \caption{Overview of all tasks in \textbf{\Ours}, Gray: inputs; Orange: outputs.}
    \label{fig:overall_tasks}
\end{figure*}

Recent foundation image and video generation models~\cite{podell2024sdxl, labs2025flux1kontextflowmatching, openai_gptimage_2025, blattmann2023stable, cui2025emu3, xie2025show, wu2025janus, chen2025janus, team2024chameleon} have made remarkable progress in semantic understanding, controllable generation, and visual reasoning through training with data on the Internet. Large-scale generative models are now capable of synthesizing highly realistic images and videos from various multimodal inputs, demonstrating impressive generalization far beyond the distribution of their training data. Such capabilities make foundation generative models an attractive starting point for embodied intelligence~\cite{yuan2026fast, ye2026world, shang2026roboscape, long2026scaling}, where robots are required to reason about complex environments and imagine future interactions before acting. 

However, embodied generation~\cite{long2025survey, liu2025aligning, li2025comprehensive} introduces challenges that differ fundamentally from conventional image and video synthesis. Unlike natural image generation, embodied scenarios require strict multi-view consistency, accurate geometric and physical coherence across cameras, explicit robot embodiment constraints, and temporally consistent interaction dynamics. The generated observations must remain compatible with robot kinematics, camera calibration, and downstream manipulation policies rather than merely appearing visually realistic. Consequently, directly applying existing foundation image or video generation models to embodied scenarios often leads to inconsistent geometry, implausible robot states, and poor compatibility with robot control. 

Recent embodied world models~\cite{agarwal2026cosmos, zhang2026qwen, shang2026roboscape} attempt to bridge this gap by continually adapting pre-trained foundation models using robot trajectories~\cite{Neill2023open_x_embodiment, tian2026interndata, yin2026geniesim30, jiang2025galaxea, walke2023bridgedata, brohan2022rt} or egocentric manipulation videos~\cite{akhtyamov2025egowalk, grauman2022ego4d}. While these approaches have demonstrated encouraging progress, their adaptation paradigm remains fundamentally different from that of foundation model pre-training. Instead of preserving the diverse image and video generation tasks that originally endow foundation models with strong semantic understanding and controllable generation, post-training is typically performed solely on robot-specific datasets. Since embodied datasets are considerably smaller, less diverse and highly repetitive compared with internet-scale visual corpora~\cite{gadre2023datacomp, schuhmann2022laion, kakaobrain2022coyo-700m}, such adaptation inevitably weakens the generalization capability inherited from foundation models and limits the diversity, controllability, and scalability of the embodied generation. 

In this work, we present \texttt{\Ours}, a Unified Embodied Synthesis Model that unifies foundation generation and embodied generation within a single training paradigm. As shown in Figure~\ref{fig:illustarte}, rather than specializing a foundation model into a robot-specific generator, \texttt{\Ours} reformulates embodied synthesis as a natural extension of foundation image and video generation. Starting from a world foundation model, \texttt{\Ours} is continually trained on both general-domain and embodied datasets under a unified autoregressive objective, jointly learning text-to-image generation, image editing, embodied scene generation, embodied transfer, and embodied video generation. 
Figure~\ref{fig:overall_tasks} exhibits the above-mentioned capabilities of \Ours.
This unified formulation preserves the rich semantic knowledge and controllable generation capability of foundation models while effectively adapting them to robot-centric multi-view reasoning and interaction modeling. 

Beyond single-step generation, \texttt{\Ours} further extends the embodied synthesis to sequential world modeling. By jointly learning interleaved embodied sequences and manipulation videos across multiple temporal resolutions, the model captures both long-horizon task progression and fine-grained interaction dynamics. More importantly, this enables embodied generation to evolve from static scene synthesis into a scalable trajectory generation engine: generated embodied scenes can be directly rolled out into temporally coherent manipulation videos, providing abundant synthetic trajectories for downstream policy learning. We believe that this capability represents an important step toward agentic embodied data generation, where foundation world models continuously generate increasingly diverse robot experiences through autonomous interaction rollout.

Extensive experiments demonstrate that \texttt{\Ours} establishes a new state of the art across both single-step and sequential embodied generation tasks. It is the first unified embodied synthesis model capable of high-quality multi-view embodied scene generation, structured controllable embodied transfer, and zero-shot embodied video generation within a single autoregressive framework. \texttt{\Ours} outperforms GPT-Image-2.0 in human evaluation on embodied scene generation and embodied transfer, achieves state-of-the-art performance on the World Arena benchmark, and significantly improves the out-of-distribution robustness of downstream robot policies using generated data. These results suggest that world foundation models can serve not only as embodied world models, but also as scalable data engines for embodied intelligence. 

Our contributions are summarized as follows: 
\begin{itemize} 

\item We introduce \textbf{\Ours}, the first unified embodied synthesis model that seamlessly integrates foundation image generation, image editing, embodied scene generation, embodied transfer, and embodied video generation within a single autoregressive framework. By jointly optimizing general-domain and embodied generation tasks under a unified continual training paradigm, \texttt{\Ours} effectively transfers the generalization capability of foundation world models to embodied generation while preserving their original visual generation ability. 

\item We propose the first unified framework for \textbf{multi-view embodied generation}, including embodied scene generation and controllable embodied transfer. By jointly training with large-scale text-to-image and image editing data, \texttt{\Ours} achieves state-of-the-art performance with superior multi-view consistency, fine-grained controllability, and instruction following, outperforming GPT-Image-2.0 in human evaluations. Furthermore, we introduce a structured control formulation that disentangles workspace, background, foreground irrelevant objects, target objects, and lighting into independent control dimensions, enabling scalable embodied video augmentation while preserving geometric consistency and interaction dynamics. The generated data substantially improves the robustness of downstream robot policies on challenging real-world manipulation tasks. 

\item We extend embodied generation from static scene synthesis to \textbf{sequential embodied world modeling} through interleaved subtask-subgoal learning and multi-FPS embodied video generation. \texttt{\Ours} achieves state-of-the-art performance on the World Arena benchmark and demonstrates strong zero-shot multi-view embodied video generation capability. By enabling coherent long-horizon rollout from synthesized embodied scenes, \texttt{\Ours} transforms foundation world models into scalable embodied data engines, providing a practical path toward agentic generation of diverse embodied interaction trajectories for future policy learning.

\end{itemize}
\section{Xiaomi-Robotics-U0}
\label{methods}

\subsection{Problem Formulation}

Our goal is to develop a unified embodied world model that supports diverse generation and prediction tasks under a single autoregressive framework. Given a multi-modal context
\begin{equation}
\mathcal{C}=\{x_1,x_2,\cdots,x_n\},
\end{equation}
where each element can be a text token, an image token, or a robot-related control token, the model learns the joint distribution over the subsequent multi-modal sequence
\begin{equation}
P(\mathcal{Y}|\mathcal{C})
=\prod_{t=1}^{T}P(y_t|y_{<t},\mathcal{C}),
\end{equation}
using standard next-token prediction (NTP).

To support embodied intelligence, \textbf{\Ours} is trained on several complementary data formats with different temporal structures. Specifically, we categorize all training samples into two groups.

\textbf{Single-step samples} describe a one-shot mapping between inputs and outputs without explicit temporal evolution. These samples include:
(i) \emph{Text-to-Image (T2I)}, where the model generates an image conditioned on textual descriptions;
(ii) \emph{Any-to-Image (X2I)}, where one to three reference images together with text are used for image editing and generation;
(iii) \emph{Embodied Scene Generation}, where the model generates the initial multi-view robot observations conditioned on robot embodiment and scene descriptions; and
(iv) \emph{Embodied Transfer}, where the model predicts transferred multi-view observations given the current embodied observations and a target scene description.

\textbf{Sequential samples} model long-horizon interactions and future evolution. We consider two representative formats:
(i) image-text interleaved subtask-subgoal sequences, where each subtask description is followed by the corresponding multi-view observations after completion; and
(ii) embodied manipulation videos collected at multiple frame rates, enabling the model to capture both sparse task-level transitions and dense interaction dynamics.

Despite their different semantics, all tasks are represented as unified multi-modal token sequences and optimized under the same autoregressive objective, allowing knowledge to be naturally shared across image generation, scene understanding, embodied reasoning, and future prediction.

\begin{figure*}[htbp]
    \centering
    \includegraphics[width=\textwidth]{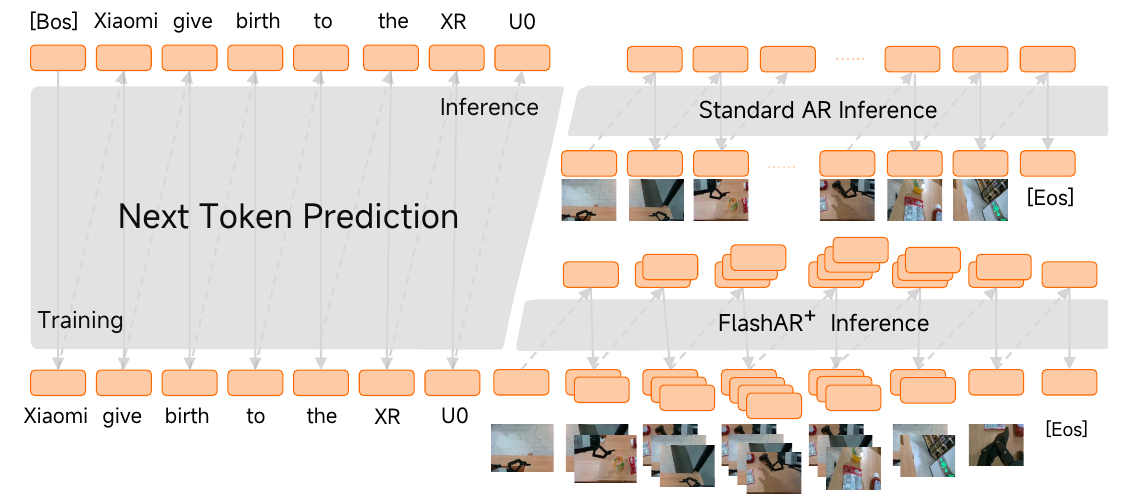}
    \caption{Overview structure of \textbf{\Ours}. The model is trained end-to-end at scale with a unified next-token prediction objective. An efficient inference extension is provided along.}
    \label{fig:structure}
\end{figure*}

\subsection{Model Architecture}


Figure~\ref{fig:structure} illustrates the overall architecture of \textbf{\Ours}. We adopt the open-sourced EMU3.5~\cite{cui2025emu3} as the initialization model, which is built upon the Qwen-3-32B decoder-only Transformer and has demonstrated strong capabilities in image generation and image-text interleaved modeling. Images are first tokenized by the IBQ Tokenizer~\cite{shi2025scalable} with a spatial compression ratio of $16\times16$. By extending the original Qwen vocabulary with the IBQ codebook, all modalities are represented within a unified discrete vocabulary and modeled using next-token prediction.

Compared with conventional embodied world models that introduce task-specific prediction heads, \\ \textbf{\Ours} formulates all generation tasks—including image generation, embodied scene synthesis, scene transfer, sequential subtask prediction, and video generation—as a single autoregressive sequence modeling problem. This unified formulation enables efficient multi-task co-training while preserving the strong visual generation capabilities inherited from the large-scale pre-trained model.

To improve inference efficiency, we further adopt an inference-accelerating adaptation inspired by FlashAR~\cite{zhou2026flasharefficientposttrainingacceleration}. Specifically, an additional vertical prediction head is introduced to decode image tokens in an anti-diagonal order, allowing multiple visual tokens to be generated simultaneously. Combined with optimized KV-cache management and continuous batching provided by vLLM~\cite{kwon2023efficient}, \textbf{\Ours} achieves up to $82.9\times$ faster image generation at 1024$\times$1024 resolution compared with the original NTP pattern, while preserving most generation quality.

\subsection{Dataset}
\begin{figure*}[!ht]
\vspace{-30pt}
    \centering
\includegraphics[width=\textwidth]{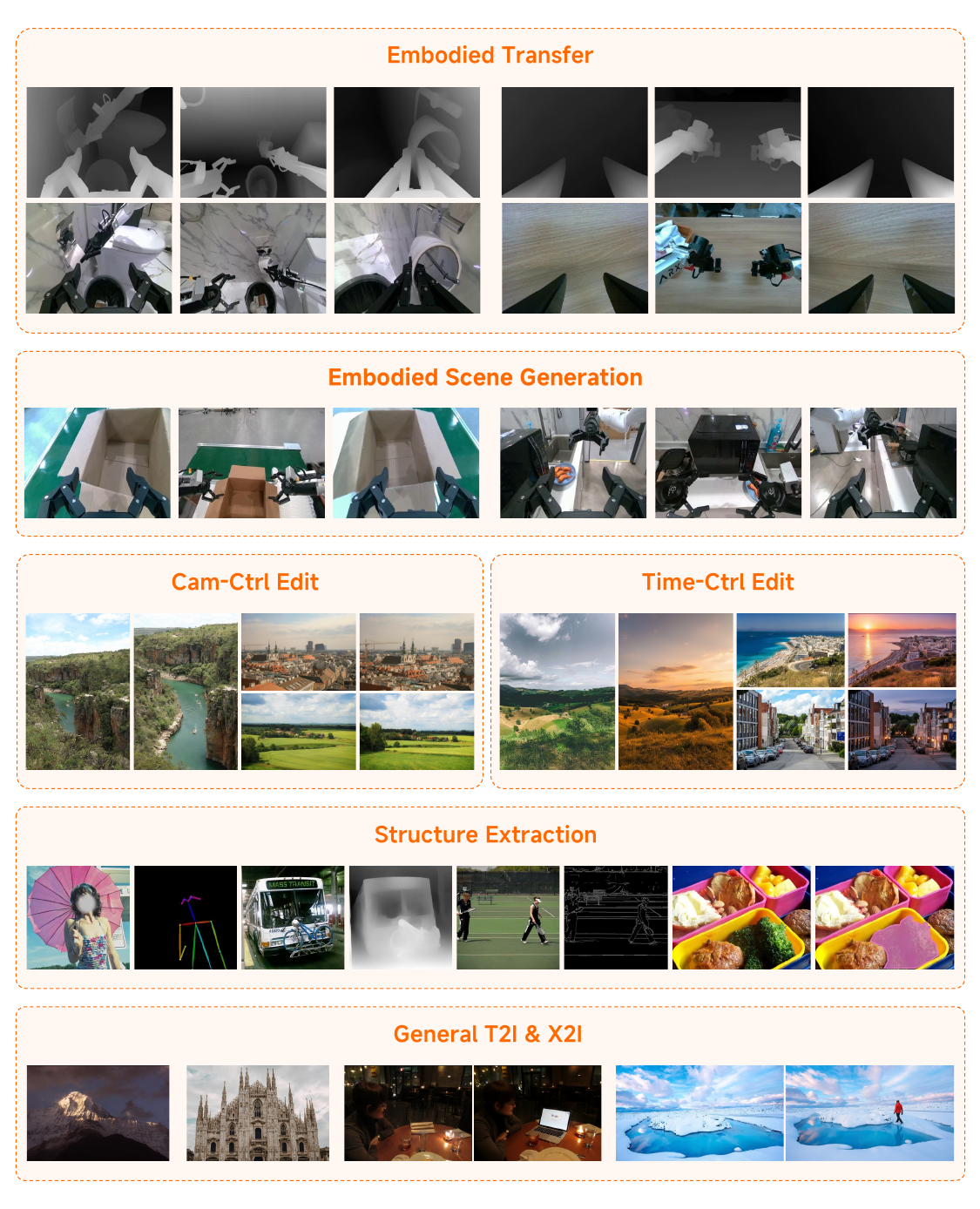}
    \caption{The illustration of the constructed single-step generation dataset. Except for the embodied transfer and the embodied scene generation data, we additionally introduce Text-to-Image(T2I) and Any-to-Image(X2I) data for co-training. Concretely speaking, we add camera-control and time-control data to improve the spatial and temporal understanding capability. For structure and geometry perception, we construct skeleton, depth, edge, and semantic mask datasets with advanced computer vision tools. To maintain the ability of the pre-trained model, we introduce general T2I and X2I datasets for co-training.}
    \label{fig:dataset_single}
    \vspace{-10pt}
\end{figure*}
\subsubsection{Dataset Curation}

\paragraph{Data Collection}
To endow \textbf{\Ours} with broad visual understanding and generation capabilities, we assemble a diverse training corpus spanning six distinct domains, each contributing complementary visual and semantic priors.

\begin{itemize}
\item \textbf{General image-text} data provides foundational visual-semantic alignment. We include ShareGPT4V~\cite{chen2024sharegpt4v} and similar large-scale vision-language benchmarks to establish a strong baseline for text-conditioned generation.

\item \textbf{Embodied manipulation} data bridges the gap between passive visual understanding and active physical interaction. This domain combines both real-world and simulated sources. On the real side, we leverage AgiBotWorld-Beta~\cite{bu2025agibot}, Open X-Embodiment~\cite{Neill2023open_x_embodiment}, and a proprietary multi-view robot manipulation dataset collected in house, which we refer to as MiBot. On the simulation side, we incorporate data from RoboTwin2.0~\cite{chen2025robotwin}, GenieSim~\cite{yin2026geniesim30}, and InternData-A1~\cite{tian2026interndata}, enabling coverage of scenarios that are difficult or dangerous to collect physically.

 \item \textbf{Autonomous driving} data captures structured outdoor scenes with complex spatial layouts and dynamic objects. We source this from Cosmos-Drive-Dreams~\cite{ren2025cosmosdrive}, which provides high-quality driving scenarios with diverse weather, lighting, and traffic conditions.

\item \textbf{Egocentric} data, drawn from EgoWalk~\cite{akhtyamov2025egowalk}, supplies first-person perspective recordings that reflect natural human viewpoint and motion patterns, enriching the model's understanding of human-centric visual experience.

\item \textbf{3D reconstruction} data offers dense multi-view correspondences and accurate geometry. We utilize ScanNet++ v2~\cite{yeshwanth2023scannet++} and DL3DV-10K~\cite{ling2024dl3dv}, both of which provide high-quality indoor and outdoor 3D scene captures.

\item \textbf{Game} data, sourced from Open-P2P~\cite{yue2026openp2p}, introduces stylized and procedurally generated environments that expand the diversity of visual distributions beyond real-world captures.
\end{itemize}

\paragraph{Data Filtering}
Raw data collected at scale inevitably contains artifacts that can impair generation quality. Inspired by Abot Physical World~\cite{chen2026abot}, we implement an autonomous data filtering pipeline including motion blur detection, image quality detection, and gripper-state alignment.

To sum up, we collect 9.5 million samples for single-step generation, which corresponds to 56.4 billion tokens. For sequential generation, we select 2.6 million video clips, which corresponds to 49.6 billion tokens.

\subsubsection{Dataset Labeling}

\paragraph{Unified Annotation Pipeline} 
Most raw samples lack fine-grained text annotations and only carry coarse task-level descriptions, which cannot support precise text-driven embodied generation. We thus construct a unified annotation pipeline to standardize supervision signals across heterogeneous data.

\begin{figure}[t]
    \centering
    \includegraphics[width=1\textwidth]{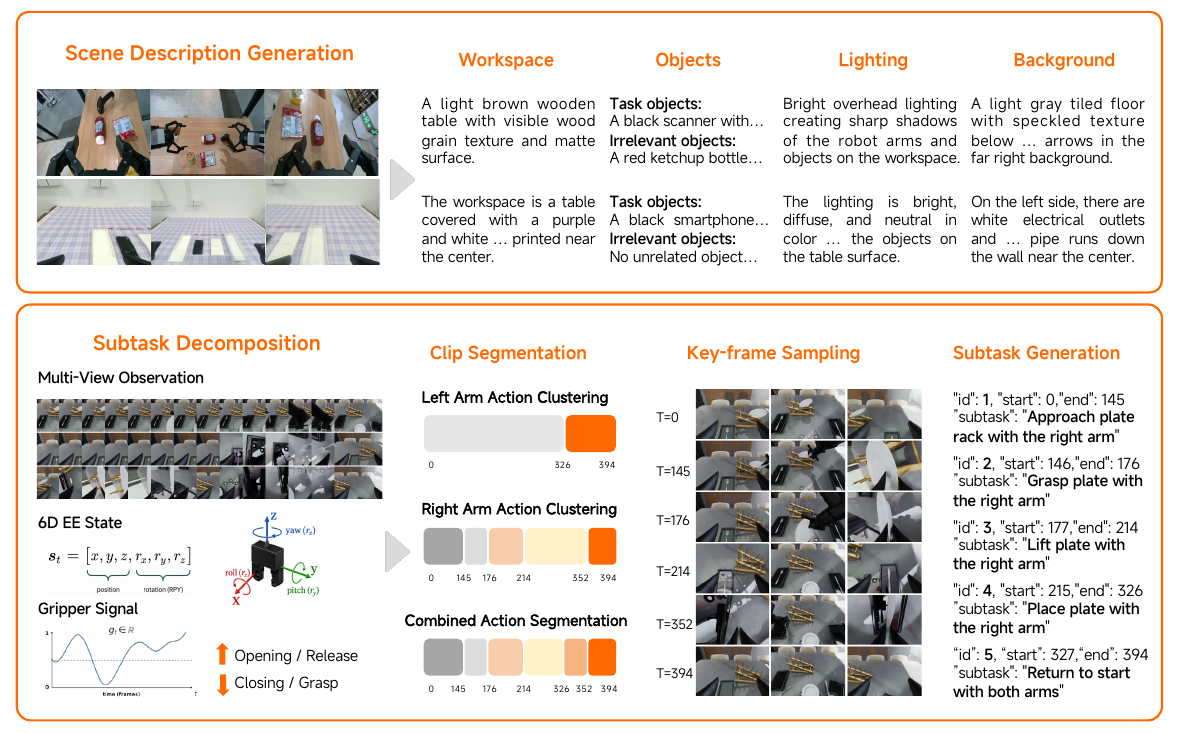}
    \caption{
Overview of the unified data labeling pipeline. 
The upper part illustrates scene description generation, where each sample is annotated with structured semantic attributes. 
The lower part shows trajectory-level subtask decomposition, including action-based clip segmentation, key-frame sampling, and subtask generation. 
This pipeline provides consistent scene, geometry, and action-level supervision across heterogeneous robotic datasets.
}
    \label{fig:data_label}
\end{figure}

\paragraph{Unified Annotation Pipeline}
We leverage Qwen3-VL-235B~\cite{bai2025qwen3}, a powerful vision-language model (VLM), to generate dense, structured annotations for generic frames, multi-source video sequences, and robotic manipulation trajectories. The pipeline consists of four decoupled branches with distinct annotation objectives:

\begin{itemize}
\item  \textbf{General Scene Annotation}
All video data spanning general visual corpora and embodied robot sequences are processed without domain restrictions. Each video clip undergoes temporal downsampling to select high-clarity keyframes. The VLM produces holistic captions describing overall scene layouts, object interactions, and cross-object spatial relationships for every sampled frame.

\item \textbf{Structured Embodied Scene Annotation}
Annotation is only performed on embodied manipulation data. Decomposed subtask outputs from trajectory processing guide scene partitioning into five orthogonal semantic dimensions: workspace, task objects, irrelevant objects, lighting, and background. Separate descriptions are generated for each dimension according to task relevance, supporting compositional reasoning and fine-grained controllable generation during inference.

\item \textbf{Geometric Modality Extraction}
Temporally consistent inverse depth maps are extracted for all frames and videos via Video Depth Anything~\cite{chen2025videodepthanything} to supply geometric conditioning signals for cross-domain embodied transfer tasks. Brighter pixels correspond to closer surfaces and darker pixels to distant regions within this depth representation, allowing the model to synthesize photorealistic RGB outputs conditioned on geometric priors.

\item \textbf{Trajectory-level Subtask Decomposition}
As illustrated in Figure~\ref{fig:data_label}, HDBSCAN clustering~\cite{McInnes2017hdbscanHD} segments robot trajectories using end-effector poses and gripper signals to identify motion and grasp transition boundaries. Keyframes with dense physical interactions are sampled at these boundaries and sent to VLM to generate temporally aligned subtask plans with frame boundaries.
Transition frames with rich physical interactions are processed with scene annotation to produce training samples for embodied scene generation and embodied transfer tasks.
\end{itemize}

\paragraph{Task-Oriented Data Organization}
To support the three training objectives of \textbf{\Ours}, we reorganize all annotated data into task-specific formats:

\begin{itemize}
    \item \textbf{Embodied Transfer.} Each training sample consists of a system instruction, a scene description, and multi-view depth maps as input, with multi-view RGB images as the supervision target. This task trains the model to hallucinate photorealistic appearance from geometric cues.
    \item \textbf{Scene Generation.} Each sample pairs a system instruction with a scene description, supervised by multi-view RGB images. This task learns text-to-multi-view-scene synthesis without geometric priors.
    \item \textbf{Video Generation.} Each sample comprises a system instruction, a task instruction, and an initial observation frame, supervised by a sequence of 15--25 RGB frames. For datasets that provide robot action labels, a sequence of robot mask images rendered from the robot action and URDF is additionally appended after the initial observation frame, serving as action guidance. This task learns to predict future visual states conditioned on task semantics and initial context.
\end{itemize}

\paragraph{Data Balancing}
\label{sec:balancing}
Our training corpus contains heterogeneous multi-domain data, which easily leads to domain bias and overfitting toward dominant data distributions. We adopt sample-level reweighting conditioned on data domains and generation task types (multi-view generation, scene editing, single-view generation), with the core target of maintaining full coverage of robotic manipulation scenarios.

Specifically, we first eliminate invalid embodied episodes, including sequences with corrupted visuals, misaligned timestamps, missing proprioceptive or action signals, failed execution records, and static trajectories lacking robot-object interactions. All remaining samples are categorized via hierarchical annotations derived from subtask and structured scene labeling. Grouping criteria cover task goals, scene contexts, target objects, spatial relations, subtask phases, and atomic manipulation skills.

Such fine-grained grouping supports more precise balancing than simple dataset-level weighting: redundant samples sharing identical embodiment-task-object combinations are downsampled, whereas clips with novel visuals, object layouts, manipulation skills or temporal stages are prioritized. Higher sampling weights are assigned to long-tail categories, including rare objects, seldom-used manipulation skills, and core interaction phases (grasping, releasing, placing, and object state transitions). The resulting training distribution eliminates redundant repeated demonstrations while retaining diverse visual scenes and comprehensive manipulation skill coverage.

\subsection{Training}


To efficiently adapt the pre-trained image generation model into a unified embodied world model while preserving its original visual generation capability, we adapt two continued training setups: \emph{single-step training} and \emph{sequential training}.

\paragraph{Single-step training} This setup focuses on aligning the model with diverse one-step generation tasks while maintaining the original image generation capability. We jointly train four complementary datasets, including Text-to-Image (T2I), Any-to-Image (X2I), Embodied Scene Generation, and Embodied Transfer.

The T2I and X2I datasets continuously reinforce the general visual generation and image editing capabilities inherited from the pre-trained model, preventing catastrophic forgetting during embodied adaptation. Meanwhile, Embodied Scene Generation teaches the model to synthesize physically plausible initial robot observations conditioned on robot embodiments and scene descriptions, while Embodied Transfer enables cross-scene multi-view generation that preserves robot interaction states and camera geometry. Co-training these tasks effectively aligns the visual generation prior with embodied scene understanding under a unified autoregressive objective.

\paragraph{Sequential training}
The sequential training extends the model to long-horizon embodied reasoning and interaction modeling.

We first introduce image-text interleaved subtask-subgoal sequences, where each language-described subtask is paired with the corresponding multi-view observations after execution. This data format teaches the model to reason over task decomposition while predicting future embodied observations, further aligning the sequence modeling capability inherited from the base model.

To enhance dynamic interaction modeling, we additionally incorporate large-scale embodied manipulation videos collected from InternData-A1~\cite{tian2026interndata}, GenieSim~\cite{yin2026geniesim30}, AgiBotWorld-Beta~\cite{bu2025agibot}, RoboCOIN~\cite{wu2025robocoin}, and RoboMIND~\cite{wu2024robomind}. Unlike conventional low-frame-rate trajectory datasets, we construct multi-frame-rate video sequences with FPS values of 1, 3, and 5. Sparse sequences encourage long-horizon planning and future prediction, whereas denser sequences enable the model to capture fine-grained manipulation dynamics and continuous interaction processes. Training across multiple temporal resolutions significantly improves the model's ability to model both high-level task progression and low-level embodied dynamics within a unified autoregressive framework.

\subsection{Inference}

To further accelerate inference, we introduce a FlashAR$^+$ extension to \textbf{\Ours}. The extension keeps the conditional modeling of reference images and other input signals unchanged, while applying parallel autoregressive generation only to the final target image region. This design is intended to preserve the multi-source conditioning capability of \textbf{\Ours} and reduce the serial decoding cost during X2I inference.

\subsubsection{FlashAR$^+$ Extension}

\begin{figure*}[htbp]
    \centering

    \begin{subfigure}[t]{0.32\textwidth}
        \centering
        \includegraphics[width=\linewidth]{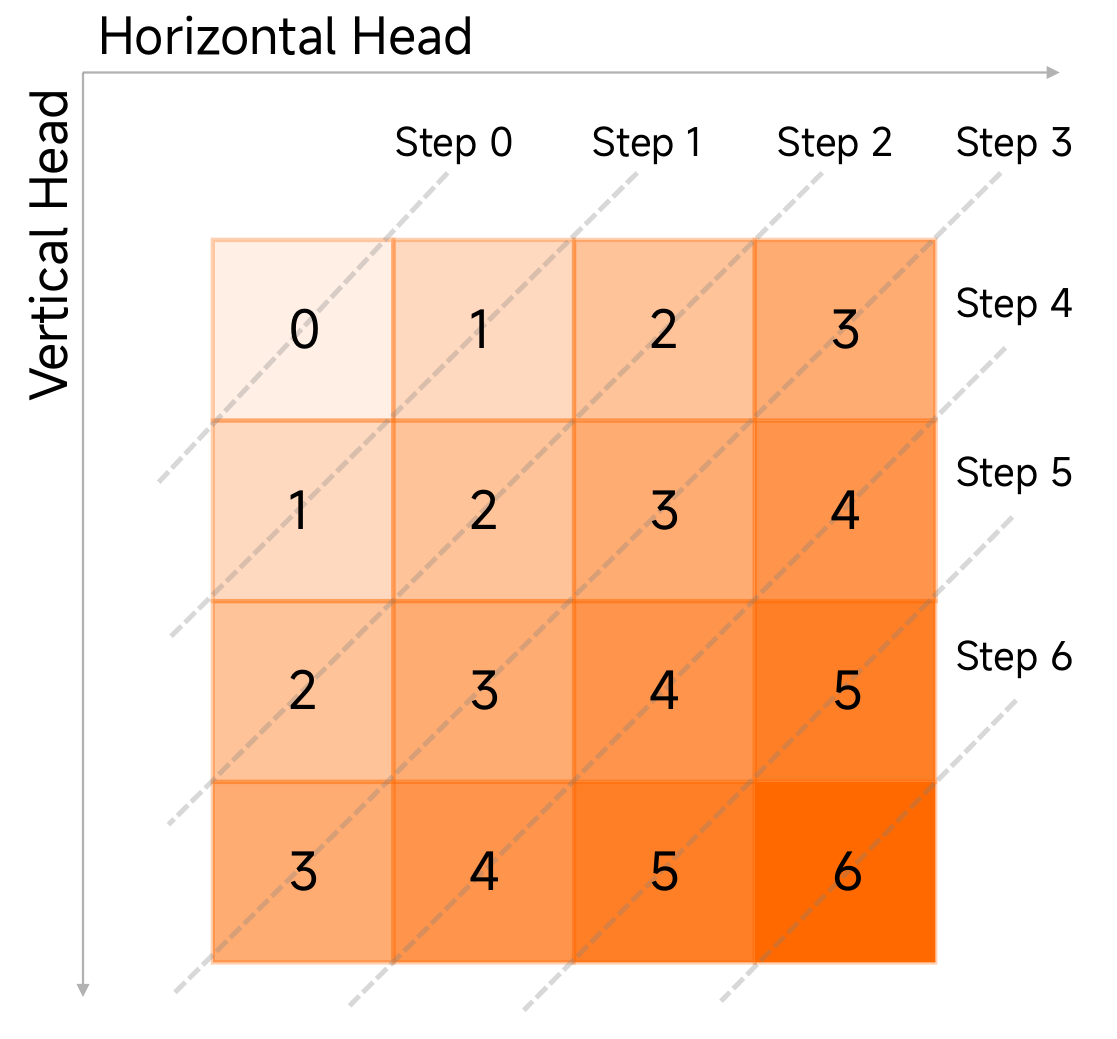}
        \caption{Diagonal step schedule.}
        \label{fig:flashar-step}
    \end{subfigure}
    \hfill
    \begin{subfigure}[t]{0.64\textwidth}
        \centering
        \includegraphics[width=\linewidth]{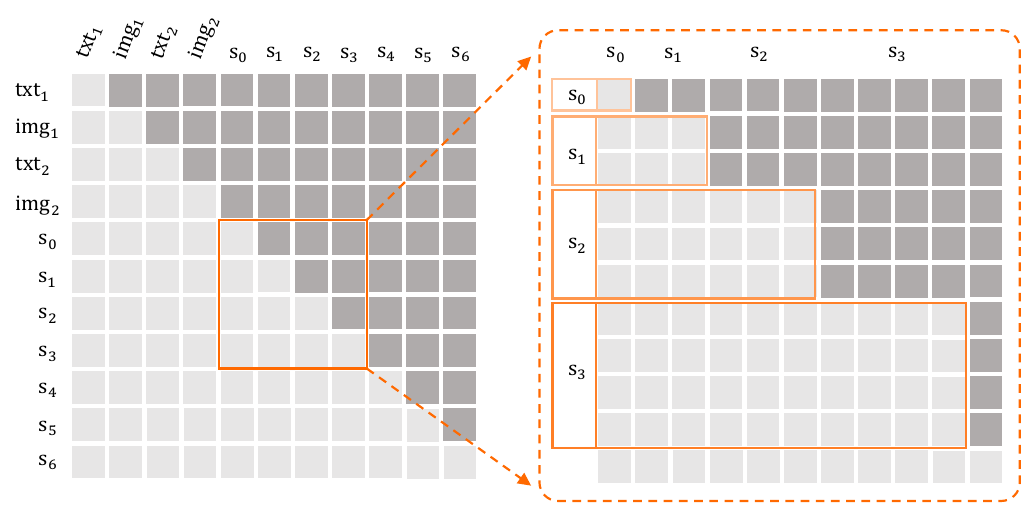}
        \caption{Prefix-conditioned step-causal mask.}
        \label{fig:flashar-mask}
    \end{subfigure}

    \caption{\textbf{\Ours} FlashAR$^+$ extension for X2I generation. Target image tokens are grouped by diagonal steps, while all input conditions are kept in the prefix and the target image is generated with a step-causal attention mask.}
    \label{fig:flashar}
\end{figure*}

\paragraph{FlashAR baseline}
FlashAR \cite{zhou2026flasharefficientposttrainingacceleration} proposes a post-training and inference acceleration paradigm for autoregressive text-to-image generation. Its core idea is to preserve the generation capability of the original model while introducing extra vertical prediction, so that image tokens can be generated in parallel along anti-diagonal groups. This reduces the number of serial decoding steps in the image region. Inspired by FlashAR, we perform FlashAR$^+$ post-training on top of \textbf{\Ours}, extending it from single-image T2I generation to broader X2I scenarios, and use the resulting model for X2I inference acceleration in \textbf{\Ours}. Figure~\ref{fig:flashar-step} demonstrates the decoding steps of the FlashAR$^+$ extension. 

In \textbf{\Ours}, X2I covers tasks such as T2I, I2I, and multi-reference image generation. The input conditions may include text instructions and one or more reference images.

\paragraph{FlashAR$^+$ mask and attention}
Compared with standard autoregressive generation, \textbf{\Ours} FlashAR$^+$ Extension introduces the key changes summarized in Table~\ref{tab:unis-flashar-extension}.

\begin{table}[h]
\centering
\small
\caption{Comparison between standard autoregressive generation and the \textbf{\Ours} FlashAR$^+$ Extension.}
\label{tab:unis-flashar-extension}
\begin{tabularx}{\textwidth}{>{\raggedright\arraybackslash}p{0.18\textwidth}
>{\raggedright\arraybackslash}p{0.25\textwidth}
>{\raggedright\arraybackslash}p{0.30\textwidth}
>{\raggedright\arraybackslash}X}
\toprule
Aspect & Standard AR & \textbf{\Ours} FlashAR$^+$ Extension & Effect \\
\midrule
Target image generation & Serial token-by-token generation in raster order & Parallel generation by anti-diagonal steps in the target image & Reduces serial decoding steps \\ \midrule
Target image attention & Standard one-dimensional causal attention & The conditional prefix is visible; the target image uses two-dimensional step-causal attention internally & Prevents information leakage \\ \midrule
Prediction heads & A single AR prediction head & Horizontal head, vertical head, and gated fusion & Supports anti-diagonal generation \\ \midrule
Directional auxiliary loss & Standard supervision & Introduces H/V loss constraints & Constrains both directional heads \\
\bottomrule
\end{tabularx}
\end{table}

During training, the FlashAR$^+$ extension identifies image blocks during data parsing and defines the target image. All preceding content, including text instructions, reference images, control conditions, and the target image header, is treated as the conditional prefix. Parallel autoregressive losses are computed only over the target image region, while reference images serve only as a conditional context. This prevents reference images from being incorrectly supervised as generation targets and ensures that the model learns to generate the target image from multi-source conditions.

The attention constraint is also restricted to the target image region, as shown in figure~\ref{fig:flashar-mask}. The conditional prefix preserves standard autoregressive visibility. Target image tokens can attend to the entire conditional prefix, but within the target image they can only attend to tokens from earlier anti-diagonal steps. For a target grid position \((r,c)\), its parallel step is defined as \(r+c\); tokens within the same step are mutually invisible, which prevents information leakage during parallel generation. Reference images, text tokens, and control tokens are not involved in the anti-diagonal mask construction inside the target image, preserving the semantic integrity of X2I conditions.

This constraint can be expressed as a visibility function within the target image. For a target-image query token \(q=(r_q,c_q)\) and key token \(k=(r_k,c_k)\),

\begin{equation}
M(q,k)={1}\left[s(r_k,c_k)<s(r_q,c_q)\right].
\end{equation}

\paragraph{FlashAR$^+$ loss} \textbf{\Ours} FlashAR$^+$ Extension introduces a key adjustment to directional auxiliary supervision. We match the horizontal auxiliary loss weight to the vertical auxiliary loss weight:

\begin{equation}
\mathcal{L}
=
\mathcal{L}_{\mathrm{fuse}}
+0.05\mathcal{L}_h
+0.05\mathcal{L}_v
+0.2\mathcal{L}_{\mathrm{distill}}.
\end{equation}

Here, \(\mathcal{L}_{\mathrm{fuse}}\) is the target-image prediction loss after H/V gated fusion, \(\mathcal{L}_h\) and \(\mathcal{L}_v\) supervise the horizontal and vertical heads respectively, and \(\mathcal{L}_{\mathrm{distill}}\) is the AR distillation term. For a target position \((r,c)\), let \(z_h(r,c)\) and \(z_v(r,c)\) denote the logits from the horizontal and vertical heads, and let \(g(r,c)\in[0,1]\) denote the gate value. The fused logits are

\begin{equation}
z_{\mathrm{fuse}}(r,c)
=
g(r,c)z_h(r,c)
\;+\;
\left(1-g(r,c)\right)z_v(r,c).
\end{equation}

The corresponding fused loss is

\begin{equation}
\mathcal{L}_{\mathrm{fuse}}
=
-\sum_{(r,c)\in\mathbf{Y}}
\log \mathrm{softmax}\left(z_{\mathrm{fuse}}(r,c)\right)_{y_{r,c}}.
\end{equation}

The directional auxiliary losses \(\mathcal{L}_h\) and \(\mathcal{L}_v\) use the same target tokens to supervise \(z_h\) and \(z_v\), respectively. The H head also needs to adapt to the target-region distribution. If \(\mathcal{L}_h=0\), the H head may fail to converge sufficiently, and the H/V gate may collapse toward the H branch in later training stages. When the gate collapses to an under-trained H head, the fused prediction degenerates into an unstable single-direction prediction, and further training can lead to degraded generation quality. Adding \(0.05\mathcal{L}_h\) explicitly supervises the H head and stabilizes H/V fusion.

\paragraph{\textbf{\Ours} FlashAR$^+$ inference}
During inference, \textbf{\Ours} first encodes the text instruction and reference images as a conditional prefix, followed by the target image header. The model then performs anti-diagonal generation only on the target image grid: positions are generated step by step in increasing order of \(r+c\), and all positions within the same step are sampled in parallel. Boundary positions use the available single-direction prediction, while interior positions use H/V gated fusion.

This procedure preserves the constraints imposed by multi-source conditions while extending image-region parallel decoding to \textbf{\Ours} FlashAR Extension inference. As a result, the model maintains Any-to-Image conditional generation capability while reducing the serial generation cost in the target image region.

\begin{figure}[t]
    \centering
    \includegraphics[width=1\textwidth]{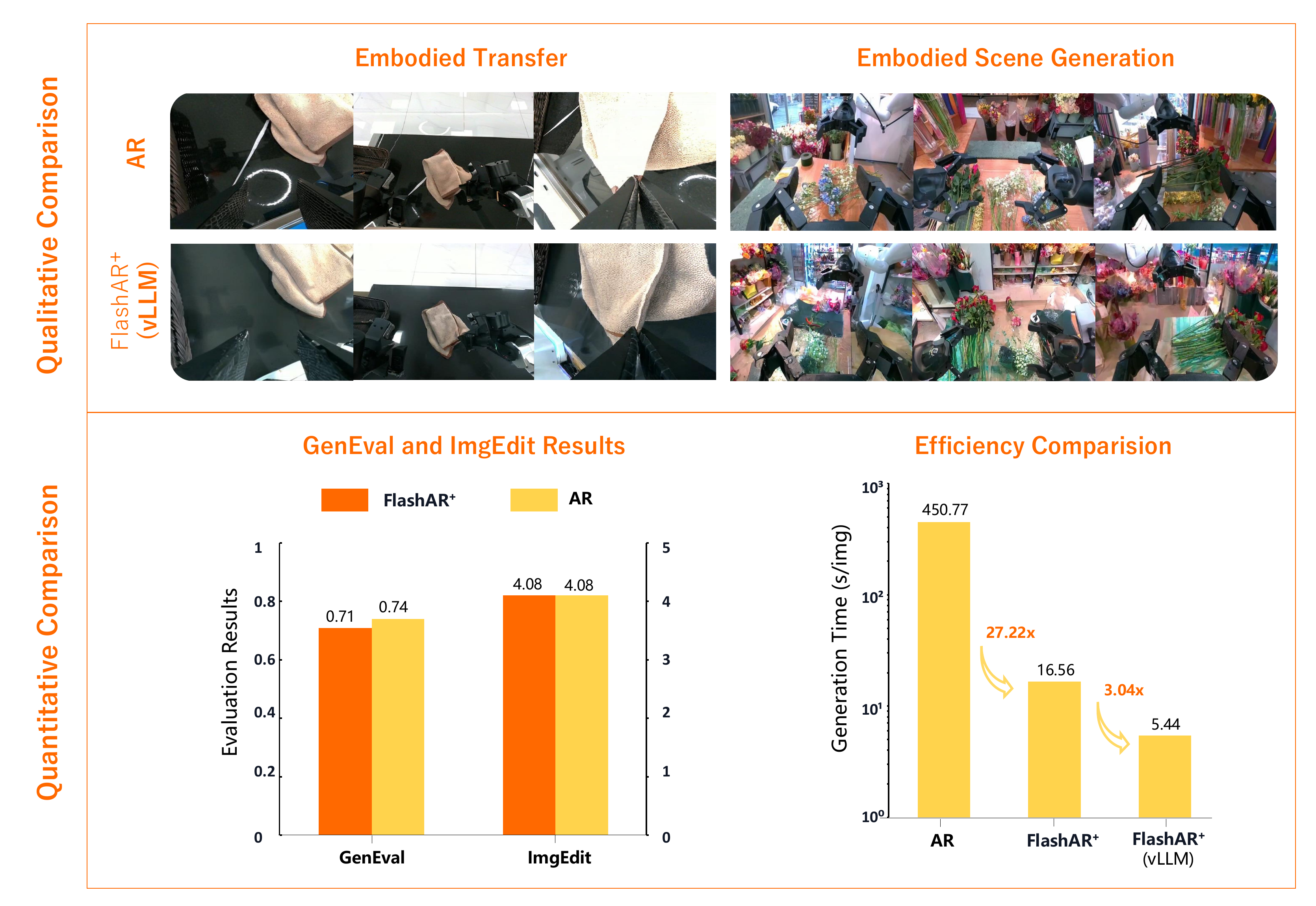}
    \caption{\textbf{Qualitative and efficiency comparison between AR, FlashAR$^{+}$, and FlashAR$^{+}$ with vLLM.} FlashAR$^{+}$ preserves high visual quality on Embodied Transfer and Embodied Scene Generation while substantially reducing generation latency.}
    \label{fig:flashar_vs_ar}
\end{figure}

\subsubsection{vLLM Integration}

To further improve the image-generation efficiency of \textbf{\Ours{}}, we integrate vLLM on top of FlashAR$^{+}$. The implementation keeps the FlashAR$^{+}$ decoding rule unchanged: vLLM handles the conditional prefix, batched execution, and paged KV-cache management, while a diagonal scheduler emits the visual tokens of each anti-diagonal step and synchronizes the paired conditional and unconditional requests used for classifier-free guidance.

As shown in Figure~\ref{fig:flashar_vs_ar}, the accelerated path preserves high generation quality on embodied tasks. On T2I generation at $1024\times1024$ resolution with a single H20 GPU, standard AR takes 450.77 seconds per image. FlashAR$^{+}$ reduces this latency to 16.56 seconds per image, achieving a $27.22\times$ speedup over AR, and FlashAR$^{+}$ with vLLM further reduces the latency to 5.44 seconds per image with \texttt{max\_num\_seq=28}, achieving an additional $3.04\times$ speedup over FlashAR$^{+}$.

\section{Experiments}
Our model unifies embodied transfer, embodied scene generation, and embodied video generation while retaining general image editing and generation capabilities. This section presents benchmark and qualitative results for our embodied transfer and scene generation tasks, which outperform GPT-Image-2.0 in human evaluation. Specifically, our embodied transfer method substantially enhances robot trajectory diversity while preserving geometric consistency, boosting the success rate of $\pi_{0.5}$ from 36.9\% to 63.2\%. We further validate the model's general image generation capacity via evaluations on the Geneval and ImageEdit benchmarks, confirming that our method retains dominant general image generation performance. For embodied video generation, our model achieves state-of-the-art results on the World Arena benchmark. It also demonstrates strong generalization to self-synthesized initial scenes, enabling a closed-loop data engine for infinite video generation conditioned on arbitrary scene descriptions and robotic instructions.

\subsection{Embodied Transfer}
\label{sec:embodied_transfer}

\paragraph{Benchmark and protocol}
To evaluate embodied transfer, we construct a benchmark of 300 samples, each consisting of a textual scene description together with a set of multi-view depth maps; the model is required to synthesize multi-view-consistent RGB images of the described robot manipulation scene.
Each sample additionally provides the multi-view RGB reference images from the original dataset. These references do not fully correspond to the (edited) scene description, but they capture the structural information beyond depth and are used to compute the benchmark metrics.
Since the depth maps are drawn from the same domain as the training set, we deliberately edit a subset of the scene descriptions so that they call for content not present in the training data, thereby probing the model's general-purpose generation ability rather than mere reconstruction.
According to how much the edited description deviates from the original scene, we split the benchmark into 150 \emph{Easy} samples (descriptions close to the underlying scene) and 150 \emph{Hard} samples (descriptions requiring substantial novel content).
We compare against GPT-Image-2, a strong image generation model.

\paragraph{Metrics}
We assess generation quality from three complementary aspects, all computed against the reference views.
For \textbf{depth consistency}, we extract monocular depth from both the generated image and the reference with a shared estimator and report SI-RMSE (scale-invariant RMSE, $\downarrow$), $\delta_1$ (the fraction of pixels with relative depth error below $1.25$, $\uparrow$), and AbsRel (absolute relative error, $\downarrow$); these measure how faithfully the generated image respects the 3D geometry imposed by the input depth.
For \textbf{structural fidelity}, we compute Canny edge F1, Precision, and Recall ($\uparrow$), capturing whether the generated layout preserves the fine-grained contours of the scene.
For \textbf{semantic grounding}, we run open-vocabulary detection and segmentation and report Segmentation mIoU and Recall ($\uparrow$), reflecting whether the described objects are correctly synthesized and localized.
As summarized in Table~\ref{tab:objective_metrics}, \texttt{\Ours} outperforms GPT-Image-2 by a large margin across all metrics on both the Easy and Hard splits, indicating markedly better adherence to the input geometry and stronger semantic alignment with the scene description.

\begin{table}[htbp]
\centering
\caption{Objective metrics on the embodied transfer benchmark. Values in parentheses denote the change of \texttt{\Ours} relative to GPT-Image-2 on the same split (signed toward the actual value; a lower value is better for $\downarrow$ metrics and a higher value for $\uparrow$ metrics).}
\label{tab:objective_metrics}
\begin{tabular}{l cc cc}
\toprule
\multirow{2}{*}{Metric} & \multicolumn{2}{c}{GPT-Image-2} & \multicolumn{2}{c}{\texttt{\Ours} (Ours)} \\
\cmidrule(lr){2-3} \cmidrule(lr){4-5}
 & Easy & Hard & Easy & Hard \\
\midrule
Depth SI-RMSE ($\downarrow$)       & 0.4007 & 0.3444 & 0.1407 (\textbf{$-$0.260}) & 0.1435 (\textbf{$-$0.201}) \\
Depth $\delta_1$ ($\uparrow$)      & 0.5307 & 0.5340 & 0.8017 (\textbf{$+$0.271}) & 0.7846 (\textbf{$+$0.251}) \\
Depth AbsRel ($\downarrow$)        & 0.4350 & 0.4045 & 0.1647 (\textbf{$-$0.270}) & 0.1702 (\textbf{$-$0.234}) \\
Canny F1 ($\uparrow$)              & 0.0519 & 0.0507 & 0.2066 (\textbf{$+$0.155}) & 0.1815 (\textbf{$+$0.131}) \\
Canny Precision ($\uparrow$)       & 0.0560 & 0.0536 & 0.2037 (\textbf{$+$0.148}) & 0.1770 (\textbf{$+$0.123}) \\
Canny Recall ($\uparrow$)          & 0.0572 & 0.0582 & 0.2290 (\textbf{$+$0.172}) & 0.2176 (\textbf{$+$0.159}) \\
Segmentation mIoU ($\uparrow$)     & 0.4105 & 0.4098 & 0.7621 (\textbf{$+$0.352}) & 0.7436 (\textbf{$+$0.334}) \\
Segmentation Recall ($\uparrow$)   & 0.6208 & 0.6524 & 0.8136 (\textbf{$+$0.193}) & 0.8032 (\textbf{$+$0.151}) \\
\bottomrule
\end{tabular}
\end{table}

\begin{figure}[!htbp]
\centering
\includegraphics[width=\linewidth]{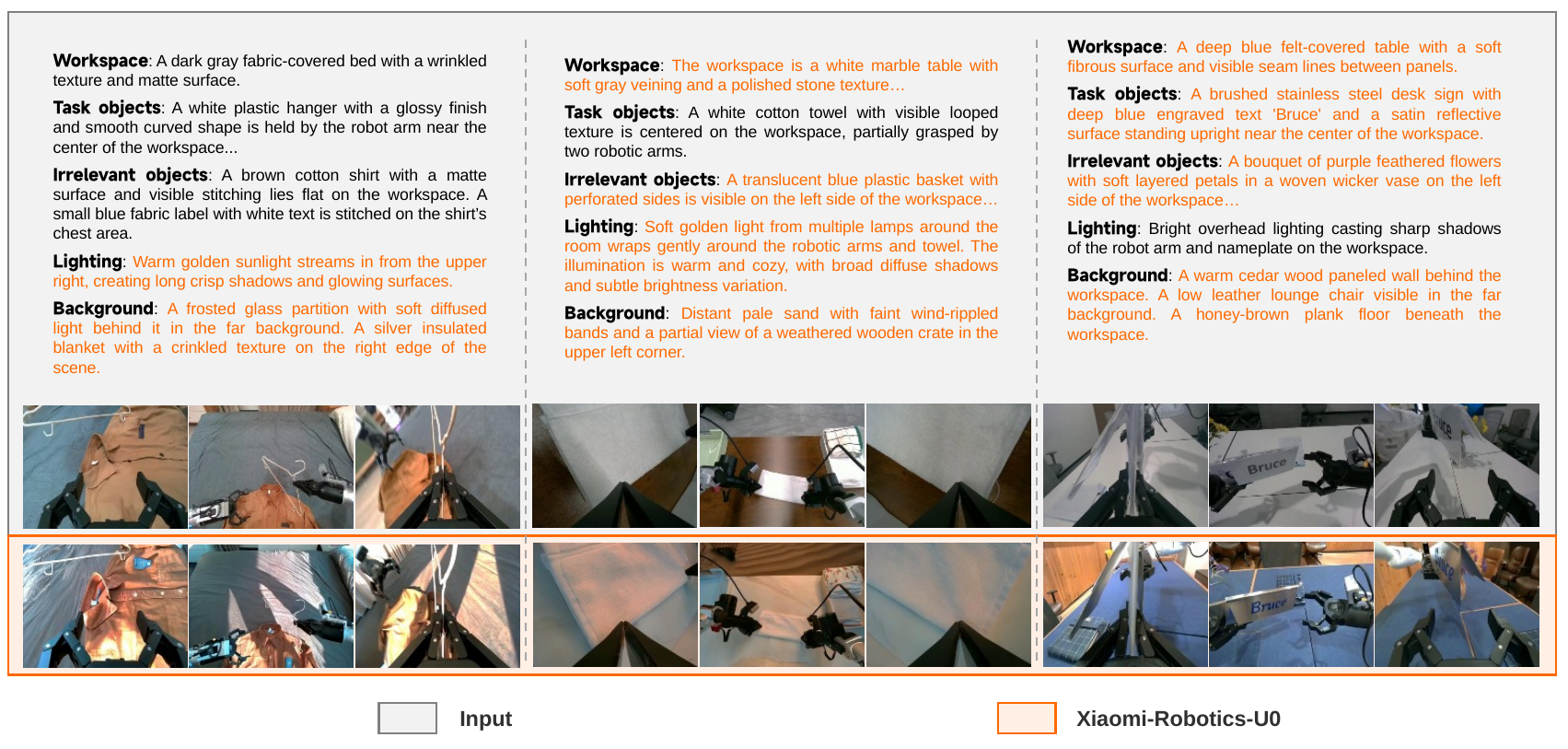}
\caption{Qualitative results on embodied transfer. For each example we show the scene description, the original multi-view RGB images from which the depth conditions are extracted, and the images generated by \texttt{\Ours}. Starting from the original images, we estimate their multi-view depth maps and edit the scene description (edited text highlighted); the depth maps together with the edited description are then fed to \texttt{\Ours}, which synthesizes multi-view-consistent RGB images that adhere to both the edited text and the input geometry.}
\label{fig:transfer}
\end{figure}

\paragraph{Qualitative results}
Figure~\ref{fig:transfer} presents qualitative examples of embodied transfer.
For each example we display the scene description, the original multi-view RGB images from the source dataset, and the images generated by \texttt{\Ours}.
Our pipeline first estimates the multi-view depth maps from the original images, then edits the scene description---with the edited spans highlighted---to call for content that departs from the original scene; the extracted depth maps and the edited description are jointly fed to the model for generation.
Across diverse scenes, and given a scene description in either natural-language or structured form, \texttt{\Ours} faithfully follows both the edited textual description and the multi-view depth conditions, correctly rendering the described objects, lighting, and background while respecting the spatial geometry of each view.
Crucially, the generated views remain mutually consistent in object identity, appearance, and arrangement, demonstrating that our unified model produces coherent multi-view scenes rather than independently plausible but conflicting images.

\begin{figure}[!htbp]
\centering
\includegraphics[width=\linewidth]{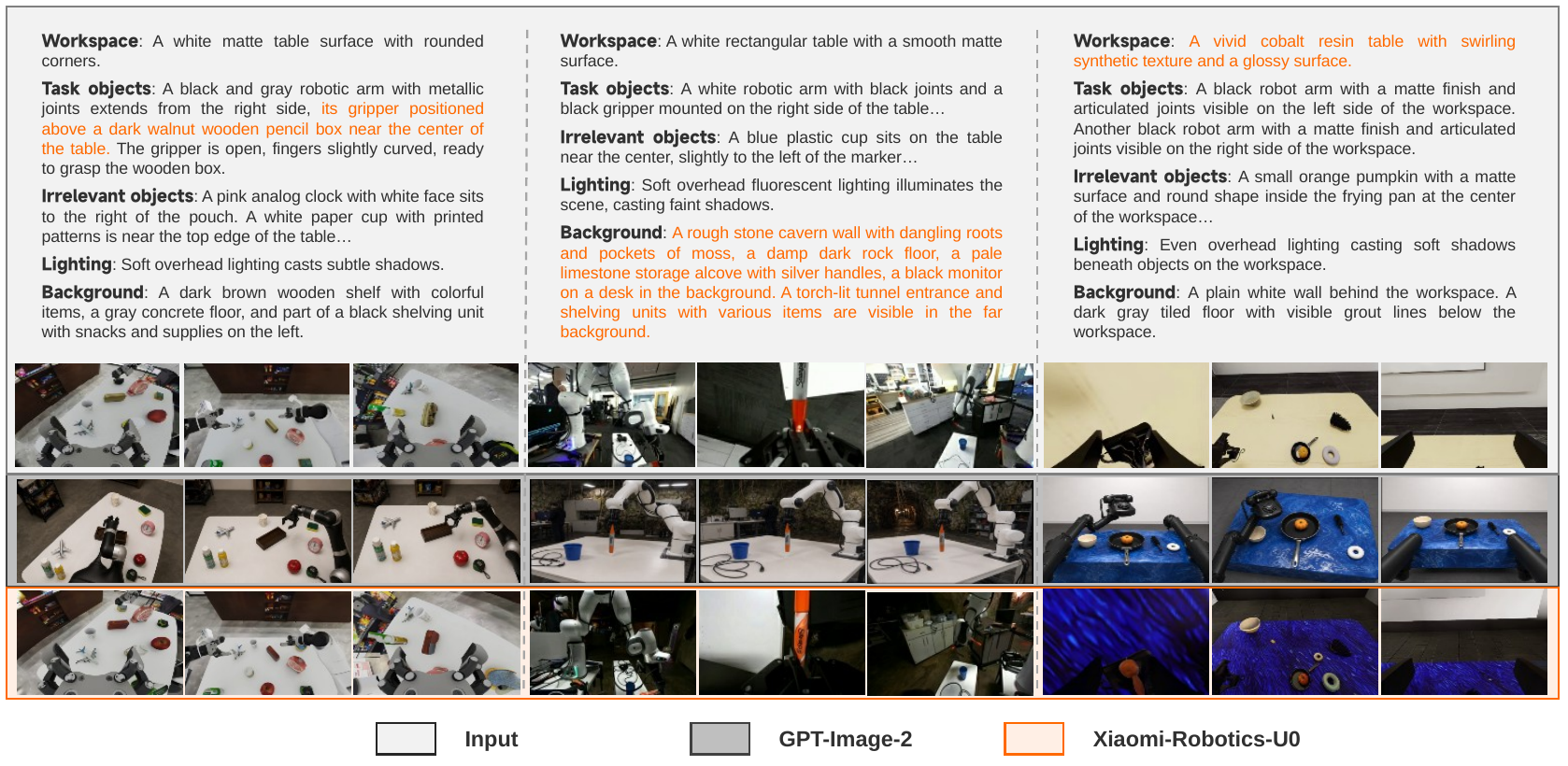}
\caption{Qualitative comparison between GPT-Image-2 and \texttt{\Ours} on the embodied transfer benchmark. GPT-Image-2 generates plausible, well-aligned images but violates the input depth and produces multi-view-inconsistent results, whereas \texttt{\Ours} adheres to the geometry of every view and keeps the views mutually consistent.}
\label{fig:transfer_compare}
\end{figure}

We further compare \texttt{\Ours} with GPT-Image-2 in Figure~\ref{fig:transfer_compare}, on three samples from the source datasets, each showing the scene description, the input multi-view depth maps, and the two models' outputs.
GPT-Image-2 produces visually appealing images with strong text--image alignment: the described objects, materials, lighting, and background are all rendered plausibly, and in isolation each view looks convincing.
However, it exhibits a poor understanding of the input geometry and of cross-view relationships.
Its outputs frequently violate the depth conditions---objects are placed at inconsistent distances, surfaces are warped, and the overall spatial layout drifts away from the imposed geometry---which is reflected in its high SI-RMSE and low $\delta_1$ in Table~\ref{tab:objective_metrics}.
Moreover, the views generated for the same scene are largely independent: object identity, pose, and arrangement fail to remain consistent across viewpoints, so the three images do not correspond to a single coherent 3D scene.
In contrast, \texttt{\Ours} faithfully respects the depth of every view while keeping the multi-view outputs mutually consistent, indicating that it not only matches the strong text-following ability of GPT-Image-2 but also genuinely grounds generation in the input geometry.

\begin{figure}[htbp]
    \centering
\includegraphics[width=0.95\textwidth]{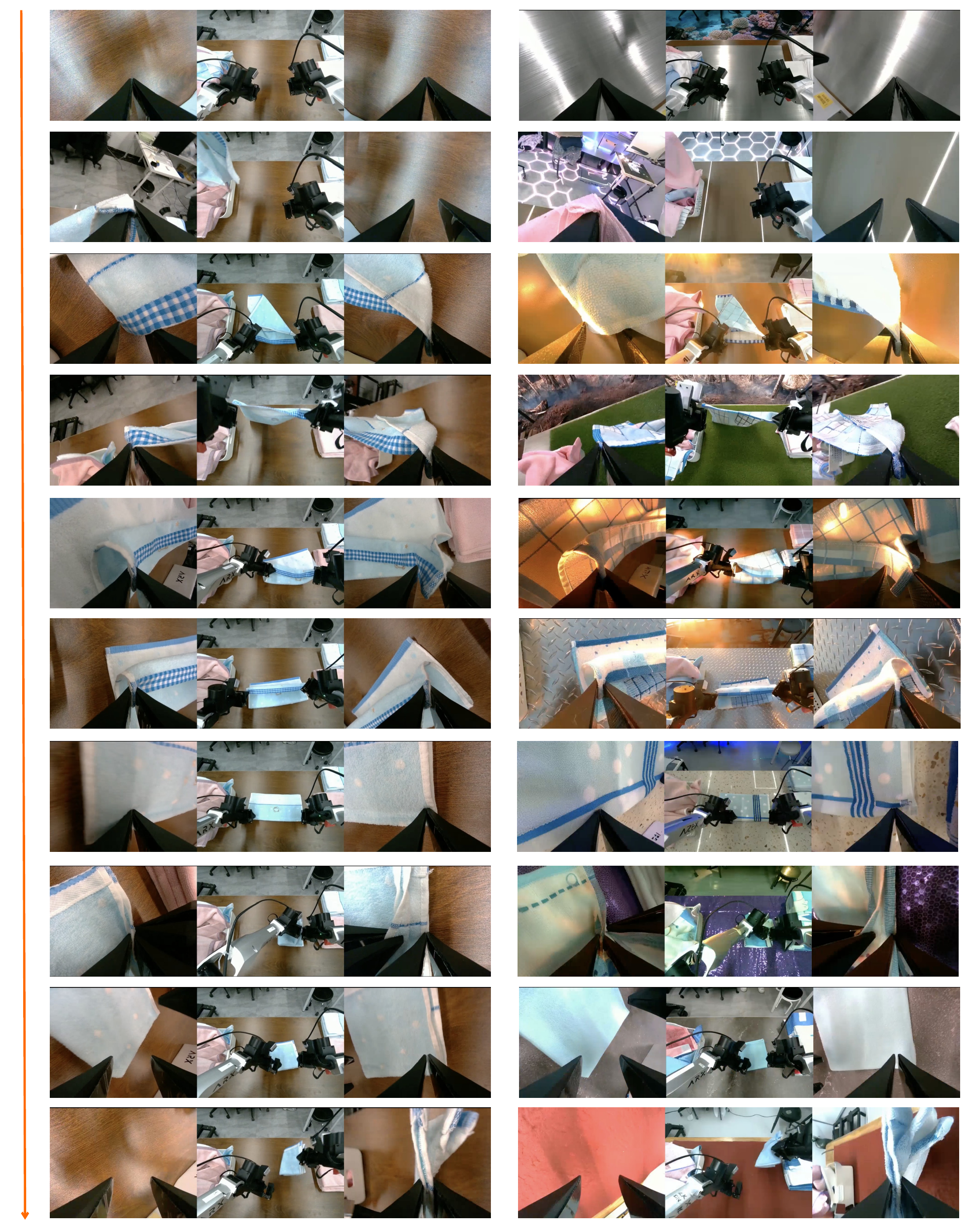}
    \caption{The zero-shot augmented keyframes of the multi-view manipulation trajectories with the embodied transfer capability of \texttt{\Ours}. We generate 1000 descriptions for the background, workspace, and lighting condition with Qwen3-VL and replace the original condition randomly. The left panel is the original keyframes, and the right corresponds to the transferred. }
    \label{fig:trans_keyframe}
\end{figure}

\begin{figure}[htbp]
    \centering
\includegraphics[width=\textwidth]{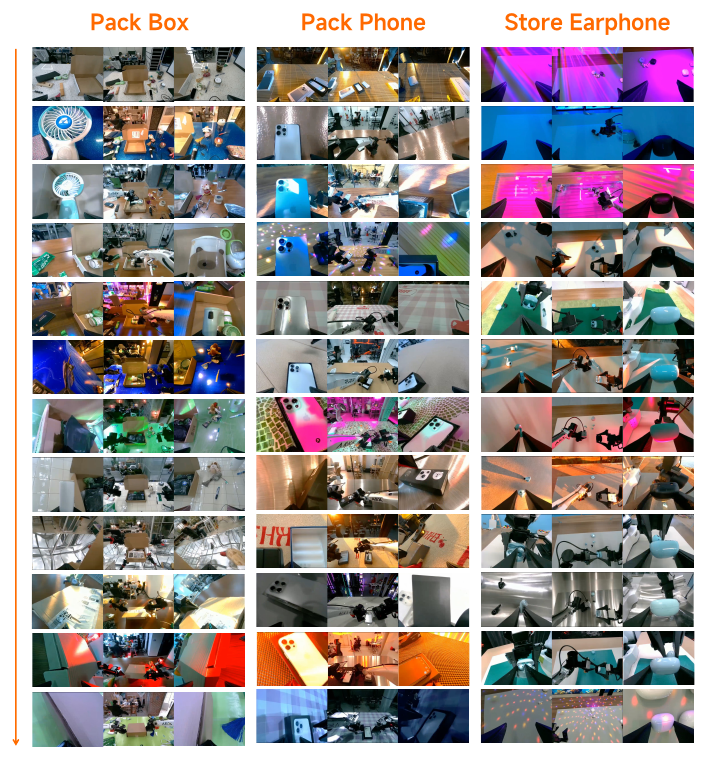}
    \caption{More results of the zero-shot augmented keyframes of the multi-view manipulation trajectories with the embodied transfer capability of \texttt{\Ours}. }
    \label{fig:more_trans_keyframe}
\end{figure}

\paragraph{Zero-shot data augmentation}
Beyond benchmark evaluation, we leverage the structured text-to-image capability of \texttt{\Ours} to augment the demonstration trajectories collected for four real-world tasks (pack box, fold towel, pack phone, and pack earphone), aiming to endow downstream policies with better generalization to novel scenes.
For each trajectory, we prompt a VLM (Qwen3-VL) to generate, conditioned on the current depth maps and the original scene description, 1000 diverse alternatives for every dimension except the task objects, encouraging maximal diversity across background, workspace, and lighting.
We then randomly combine these per-dimension descriptions and feed them to \texttt{\Ours} for generation.
As shown in Figure~\ref{fig:trans_keyframe} and Figure~\ref{fig:more_trans_keyframe}, the resulting scenes exhibit rich diversity, and none of these compositions appear in the training set, demonstrating that \texttt{\Ours} can substantially expand the visual coverage of the collected trajectories in a zero-shot manner.

\subsection{Embodied Scene Generation}
\label{sec:embodied_scene_gen}
\begin{figure}[htbp]
    \centering
    \includegraphics[width=1\textwidth]{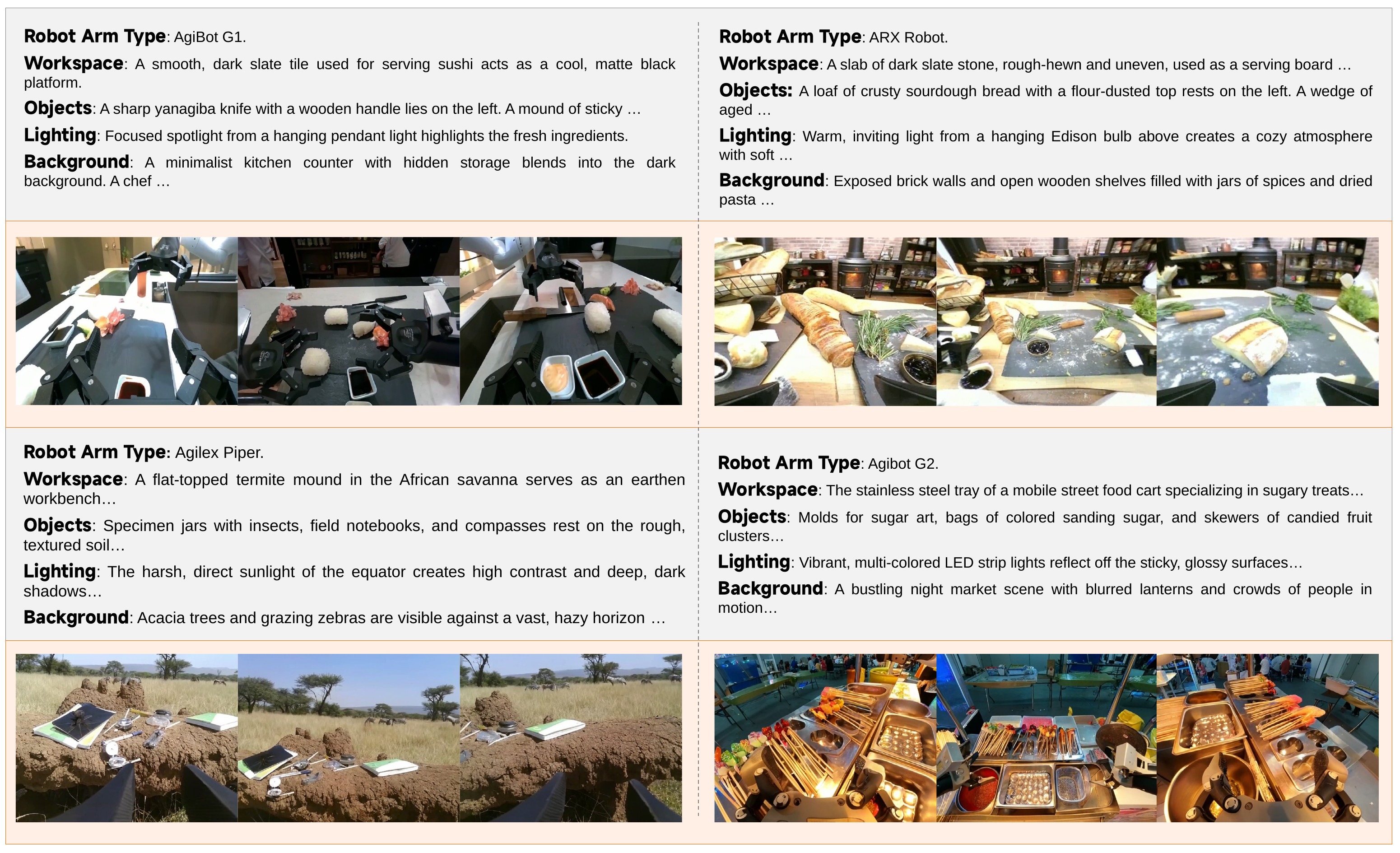}
    \caption{Qualitative results on embodied scene generation.}
    \label{fig:init_scene_results}
\end{figure}

\begin{figure}[!t]
    \centering
    \includegraphics[width=1\linewidth]{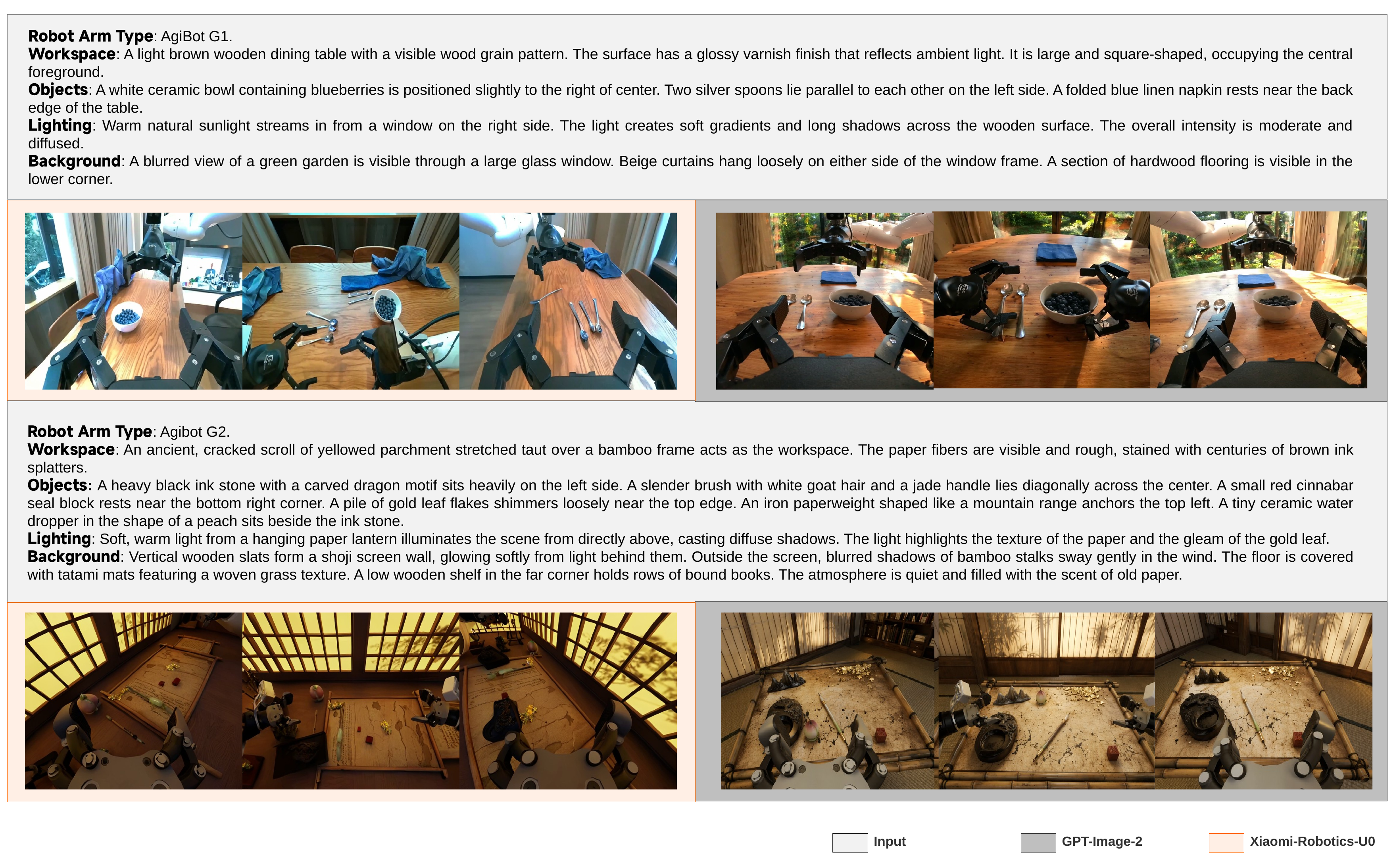}
    \caption{Qualitative comparison between GPT-Image-2 and \texttt{\Ours} on the embodied scence generation benchmark.}
    \label{fig:scene_gen_compare}
\end{figure}

\begin{figure}[!h]
    \centering
    \includegraphics[width=0.9\linewidth]{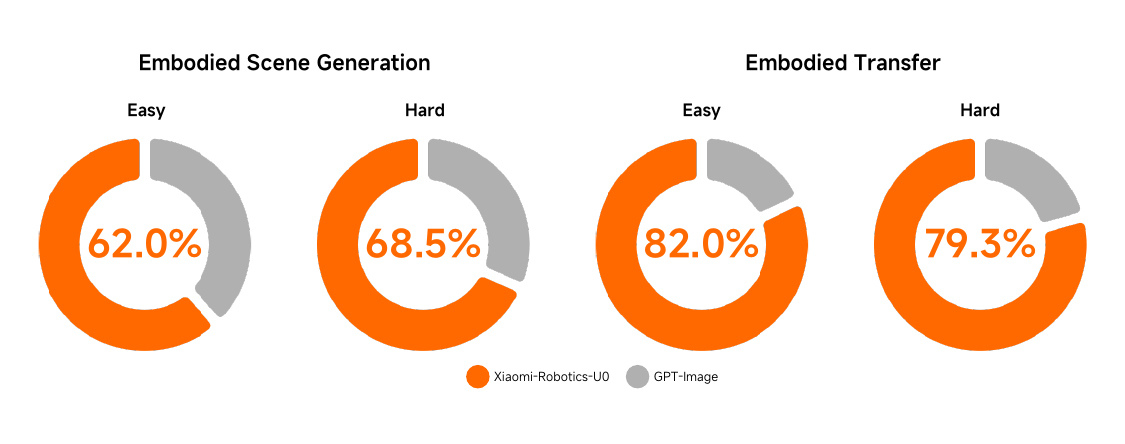}
    \vspace{-10pt}
    \caption{Pairwise evaluation results of embodied transfer and embodied scene generation.}
    \label{fig:pair_compare}
\end{figure}

\paragraph{Benchmark and Protocol}
To evaluate embodied scene generation, we construct a dedicated benchmark using LLM-generated scene prompts.
The benchmark contains 400 samples split equally into two difficulty levels: 200 Easy samples for conventional indoor manipulation scenes, and 200 Hard samples for unconstrained open-domain scenes.
Each prompt specifies a robot arm type and a structured scene description covering workspace layout, object configuration, lighting, and background.
We compare \texttt{\Ours} against GPT-Image-2 as the baseline.
To ensure a fair comparison, multi-view reference images corresponding to each target robot arm type are provided to GPT-Image-2 as visual context.

\paragraph{Human-Based Pairwise Evaluation}
Quantifying cross-view geometric consistency remains an open challenge, as no automated metric reliably evaluates global 3D coherence across viewpoints.
We thus adopt a human pairwise preference protocol for quantitative assessment.
Annotators are presented with paired outputs from \texttt{\Ours} and GPT-Image-2 in randomly shuffled left-right orders to mitigate positional preference bias. Annotators select the superior result based on two criteria: (1) multi-view geometric consistency, measuring unified 3D layouts with invariant object positions, scales, and occlusion across views;
(2) textual instruction following, measuring alignment with the workspace, objects, lighting, and background in prompts.

Figure~\ref{fig:pair_compare} reports the overall win rate. \texttt{\Ours} outperforms GPT-Image-2 in most comparisons, achieving clear advantages in multi-view consistency while maintaining comparable instruction-following performance.
Qualitative comparisons in Figure~\ref{fig:scene_gen_compare} show that GPT-Image-2 produces high-quality images but suffers from severe cross-view geometric inconsistencies, with conflicting object positions, scales, and layouts across cameras. By contrast, \texttt{\Ours} generates geometrically coherent multi-view observations with consistent spatial layouts, while retaining photorealistic quality. This verifies that \texttt{\Ours} simultaneously achieves high visual fidelity and robust cross-view consistency.

\paragraph{Qualitative Results}
Figure~\ref{fig:init_scene_results} presents qualitative results covering four robot arm types. \texttt{\Ours} generates photorealistic and geometrically consistent multi-view initial scenes for diverse scene prompts, demonstrating strong open-ended generalization to arbitrary textual descriptions. This shows that our model extends conventional single-view text-to-image synthesis to multi-view robotic observation generation by learning intrinsic geometric relations between arm configurations and camera projection. Notably, among the four evaluated robot arm types, the AGIBOT G2 is trained exclusively on simulated data, yet \texttt{\Ours} can still produce photorealistic, real-world-aligned observations for this specific arm type. This validates the model’s promising sim-to-real generalization potential via learned real-world visual priors.

\begin{figure}[t]
    \centering
    \includegraphics[width=\linewidth]{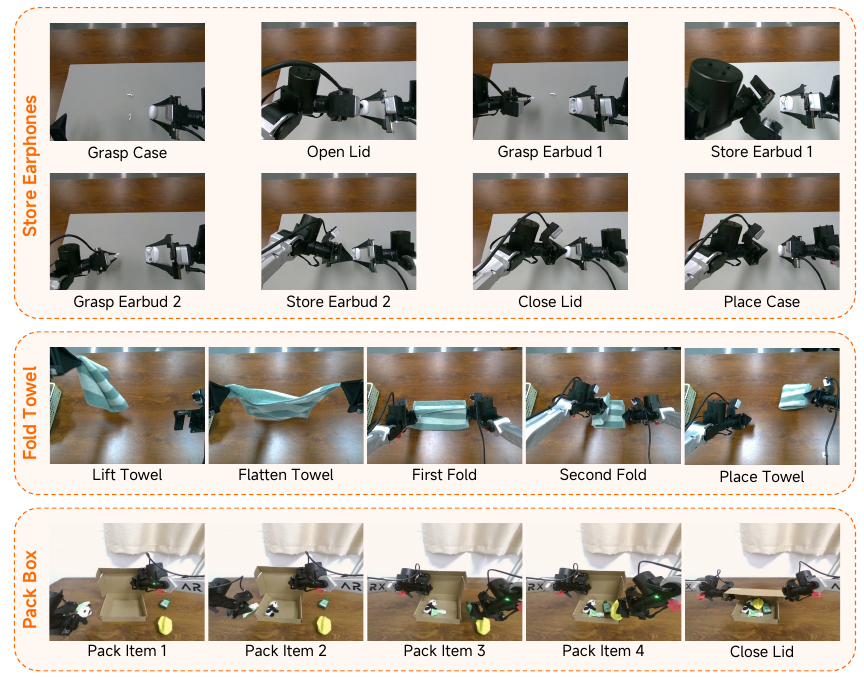}
    \caption{Stage-level progress visualization for each task. Each row corresponds to one task, and each column shows the head view at the completion of a progress milestone.
    For \textit{Store Earphones}, the eight milestones run from grasping the case, through storing both earbuds, to closing the lid and placing the case. For \textit{Fold Towel}, the milestones cover lifting and flattening the towel, two sequential folds, and placing the folded towel. For \textit{Pack Box}, they cover four pick-and-place steps followed by closing the box lid.}
    \label{fig:task_stages}
\end{figure}

\subsection{Real World Experiments}

\label{sec:real_world_experiments}

We conduct the real-world evaluation to investigate whether the proposed world model improves downstream policy learning. Specifically, we use the \texttt{\Ours} model to augment the expert demonstrations, and compare the VLA policies trained on the original and augmented data across a suite of real-world tasks.

\paragraph{Task suite} The benchmark consists of three real-world tabletop manipulation tasks: \textit{Store Earphones}, \textit{Fold Towel}, and \textit{Pack Box}, encompassing diverse challenges such as dexterous manipulation, deformable object handling, and multi-object interaction. Each task is defined by a language instruction and an ordered set of progress milestones.
Figure~\ref{fig:task_stages} visualizes the milestone states for each task, providing a concrete reference for the progress metric defined later in this section.

\newcommand{\SnapshotCell}[1]{%
    \raisebox{-0.3em}{\includegraphics[width=\linewidth,keepaspectratio]{#1}}%
}

\begin{figure}[!t]
    \centering
    \includegraphics[width=\linewidth]{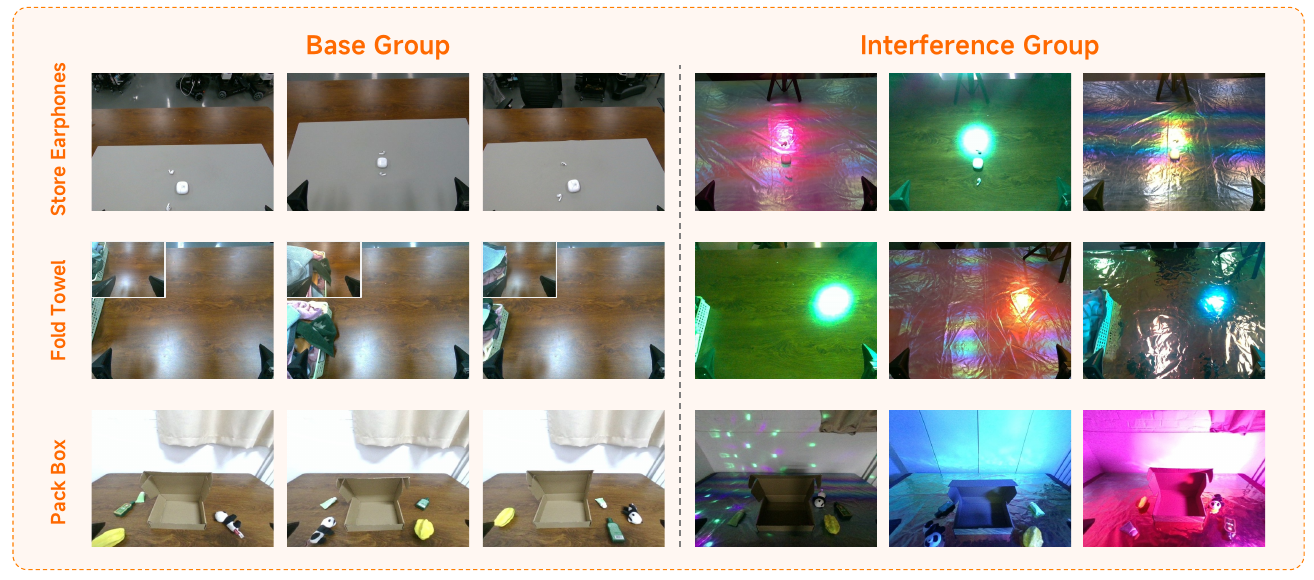}
    \caption{
    Task layouts for all three tasks.
      The base group varies the object layout for each task.
      For \textit{Store Earphones}, different layouts vary the positions of the case and the earbuds.
      For \textit{Fold Towel}, the towel is initialized in a randomized configuration, which is clearer in the left-wrist view shown as an inset at the top-left corner.
      For \textit{Pack Box}, a new layout is generated by randomly placing the four objects on the table.
      The interference group uses the same layout randomization but additionally introduces substantial changes in tablecloth and lighting.
    }
    \label{fig:layout}
\end{figure}

\begin{figure}[t]
    \centering
    \includegraphics[width=\linewidth]{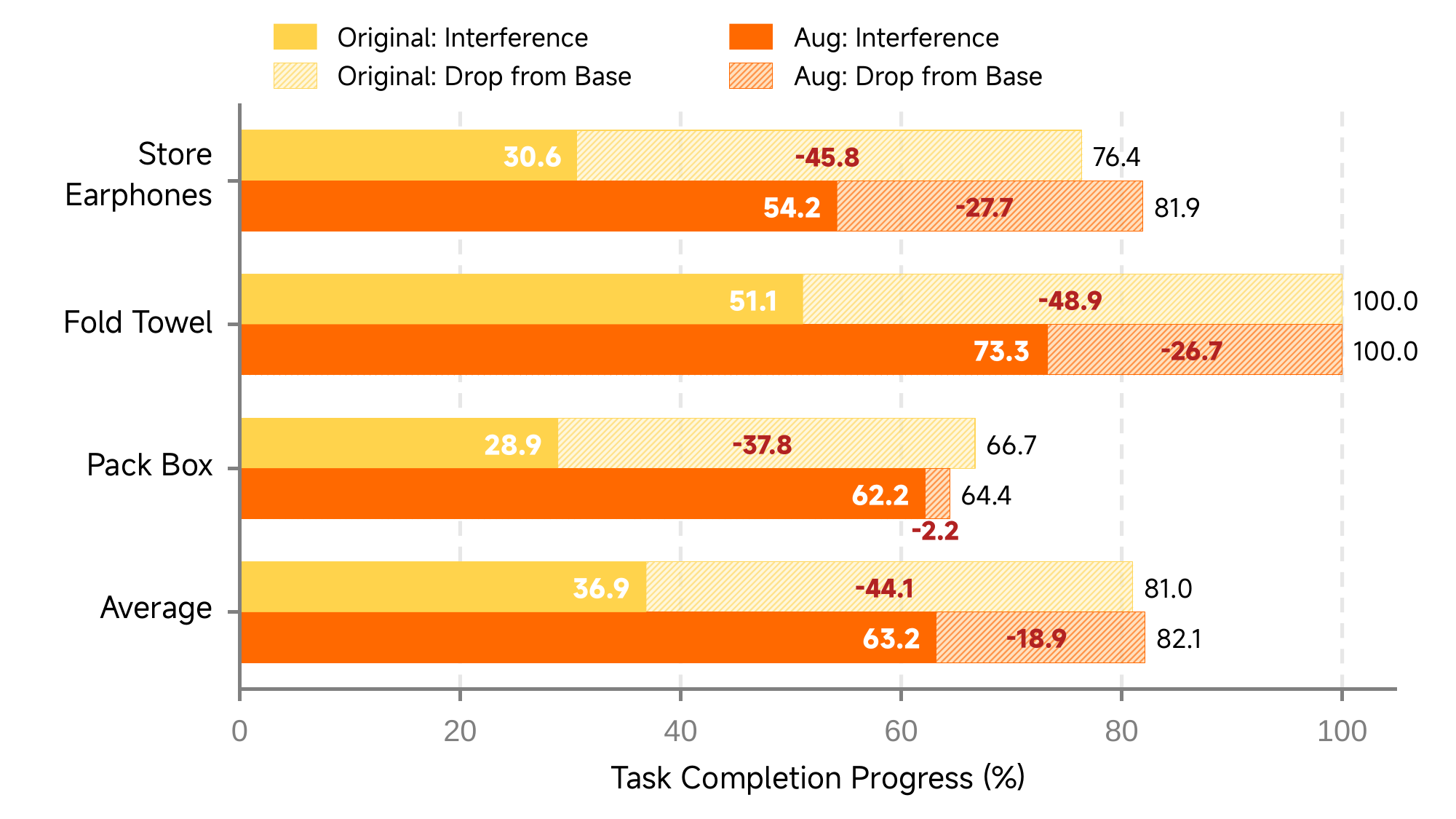}
    \caption{Comparison of task completion progress between Original and \texttt{\Ours}-Aug, split by evaluation group. 
    \texttt{\Ours}-Aug is on par with Original under in-distribution conditions (base group) but achieves substantially higher progress under held-out backgrounds and lighting (interference group), leading to a clear overall advantage.}
    \label{fig:real_world_results}
\end{figure}

\paragraph{Policy training and data}
For each task in the suite above, we independently post-train a $\pi_{0.5}$~\cite{intelligence2025pi_0_5} policy starting from the  \texttt{pi05\_base} checkpoint, using its official PyTorch implementation.
All training settings are kept identical across compared policies; the only variable is the supervised fine-tuning data mixture.
For each task, we collect approximately 40 hours of real-world demonstrations on the robot platform described below, forming the task-specific clean data.
To construct augmentation data, we apply the embodied scene generation pipeline (Section~\ref{sec:embodied_scene_gen}) to the clean demonstrations, producing approximately 40 hours of style-transferred episodes per task that vary visual appearance---backgrounds, lighting, and textures---while preserving the original robot states and action labels.
We compare two resulting per-task policies:
\textbf{Original} is post-trained exclusively on the task-specific clean data;
\textbf{\texttt{\Ours}-Aug} is post-trained on the mixture of clean data and its corresponding style-transferred augmentation.
No inference-time world-model module is used during robot execution; the world model influences the policy solely through the post-training data mixture.

\paragraph{Evaluation schedule}
To evaluate robustness beyond the original demonstration distribution while keeping the comparison controlled, we evaluate each policy under two groups of real-world conditions: a base group and an interference group.
Each group contains three object-layout variants per task, and both policies are evaluated on an identical condition schedule.
In the base group, the three layouts differ only in object placement; the background uses tablecloths seen in the demonstration data, and the scene is evaluated under nominal laboratory illumination.
In the interference group, the layouts again vary object placement, but the background is replaced by held-out tablecloths absent from the demonstration data---often with more complex patterns and different textures or surface reflectance. The interference group further introduces held-out illumination, including low light, colored light, and dynamic patterned lighting such as disco-ball effects. The full set of evaluation layouts for every task and group is shown in Figure~\ref{fig:layout}.
For every policy, task, test group, and layout, we run three independent real-robot trials.
This yields $2\times3\times3=18$ trials (two groups $\times$ three layouts $\times$ three trials) per policy-task pair.

\paragraph{Robot setup}
We deploy all trained policies on a bimanual ARX robot platform with two 6-DoF arms, each equipped with a 1-DoF parallel-jaw gripper.
At each policy query, the observation consists of the current proprioceptive arm state and three RGB views: one head view from the head camera and two wrist views from the left and right wrist cameras. For \textit{Store Earphones} and \textit{Fold Towel}, these views are captured with Intel RealSense D405 cameras; for \textit{Pack Box}, we use fisheye cameras in the same viewpoint configuration to cover the larger packing workspace. The fisheye images are undistorted before use, both as policy inputs and as inputs to the \texttt{\Ours} augmentation.
During real-robot evaluation, the policy is running on a separate inference server and communicates with the robot control process over a WebSocket connection on the local-area network.
At 1\,Hz, the robot transmits the current arm state and the three RGB views to the server, which returns an action chunk that is executed until the next query.
The observation stream, communication stack, and action-chunk execution interface are identical across all compared policies.

\paragraph{Metric}
We use task completion progress as our sole evaluation metric. Rather than scoring only full success, progress assigns partial credit based on the ordered milestones a rollout completes, making it better suited to long-horizon manipulation, where a policy may reliably solve early subgoals yet fail at a later stage. Specifically, if a task has $K$ ordered milestones and the rollout completes the first $\ell$ of them, its progress is $\ell/K\in[0,1]$, where $1$ denotes full success. For task $t$ and group $g\in\{\text{base},\text{interference}\}$, the reported progress of the tested policy is defined as:
\begin{equation}
\mathrm{Prog}(t,g)=
\frac{1}{N}
\sum_{i=1}^{N}
\frac{\ell_{i}}{K_t},
\end{equation}
where $N$ is the number of rollouts for each task and group (three layouts $\times$ three trials in our setting). A higher progress score for \texttt{\Ours}-Aug indicates that the world model provides useful post-training variation for improving real-world robustness across both fully and partially completed executions.

\paragraph{Experiment results}
Figure~\ref{fig:real_world_results} reports the task completion progress of both policies.
In the base group, where test conditions closely resemble the demonstration distribution, \texttt{\Ours}-Aug performs comparably to Original, with a slight decrease on some tasks---an expected consequence of sharing training capacity with augmented data that does not add new information for in-distribution scenes.
In the interference group, however, \texttt{\Ours}-Aug consistently outperforms Original across all tasks, indicating that style-transferred augmentation induces visual invariance to novel backgrounds and lighting unseen in the clean demonstrations.
Overall, the robustness gains under distribution shift outweigh the marginal in-distribution trade-off, which is especially valuable for real-world deployment, where scene conditions cannot be tightly controlled.

\begin{figure}[!t]
    \centering
    \includegraphics[width=\linewidth]{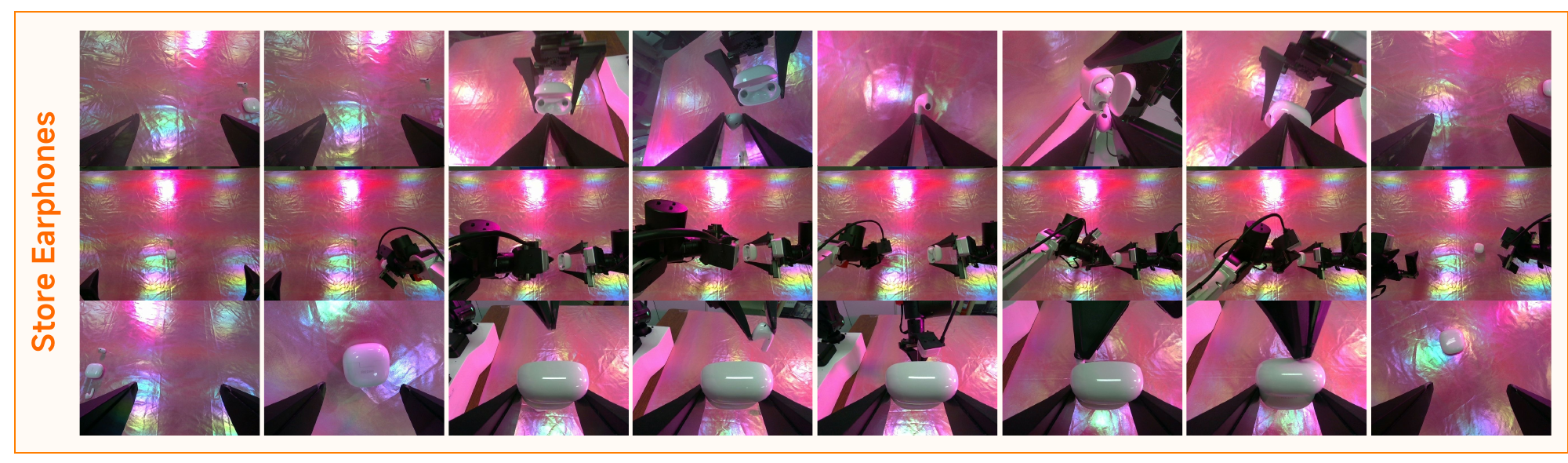}\\
    \includegraphics[width=\linewidth]{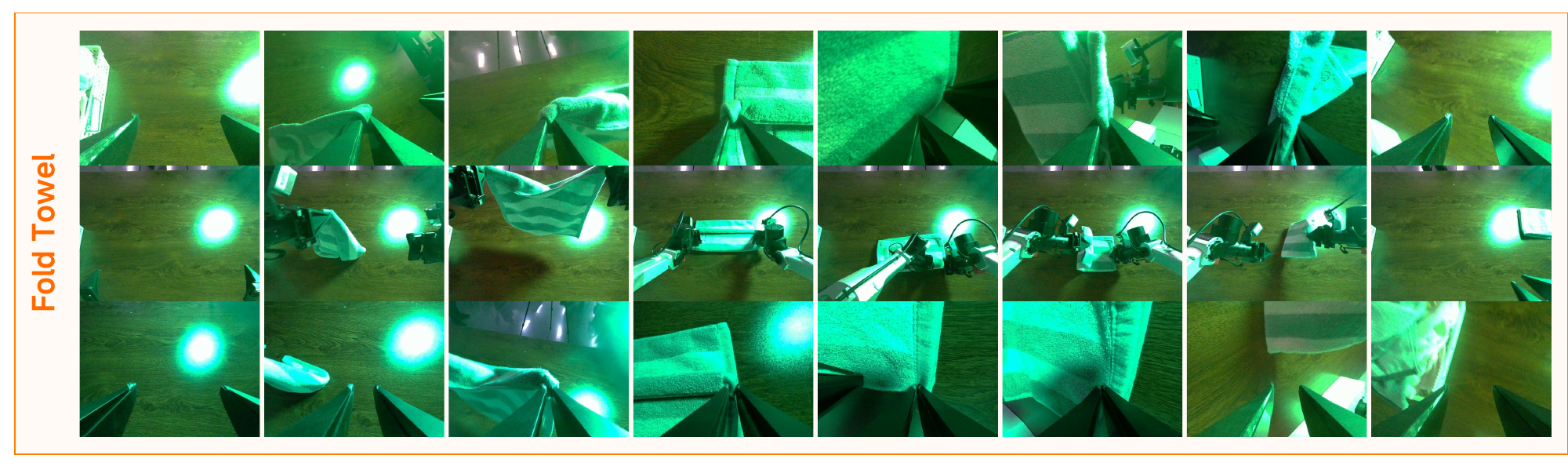}\\
    \includegraphics[width=\linewidth]{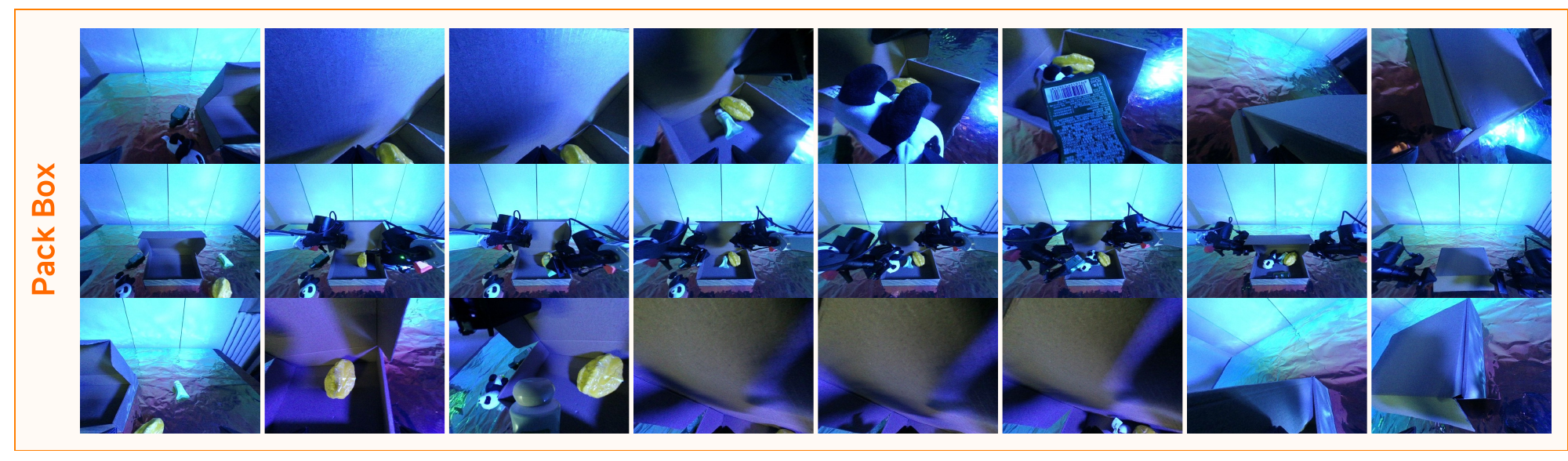}
    \caption{\texttt{\Ours}-Aug rollouts under interference scenarios for \textit{Store Earphones}, \textit{Fold Towel}, and \textit{Pack Box} (top to bottom). Within each task, the three rows are the head view, left wrist view, and right wrist view, with frames sampled across the trajectory.}
    \label{fig:rollout_episodes}
\end{figure}

To illustrate this gap qualitatively, Figure~\ref{fig:rollout_episodes} presents representative \texttt{\Ours}-Aug rollouts under interference conditions for all three tasks.
Despite challenging perturbations---held-out tablecloths, colored lighting, and projected patterns---the \texttt{\Ours}-Aug policy still executes the intended manipulation sequences:
it stores both earbuds and closes the case in \textit{Store Earphones}, completes the full folding sequence in \textit{Fold Towel}, and packs all four objects before closing the lid in \textit{Pack Box}.

\begin{figure}[!t]
    \centering
    \includegraphics[width=0.8\linewidth]{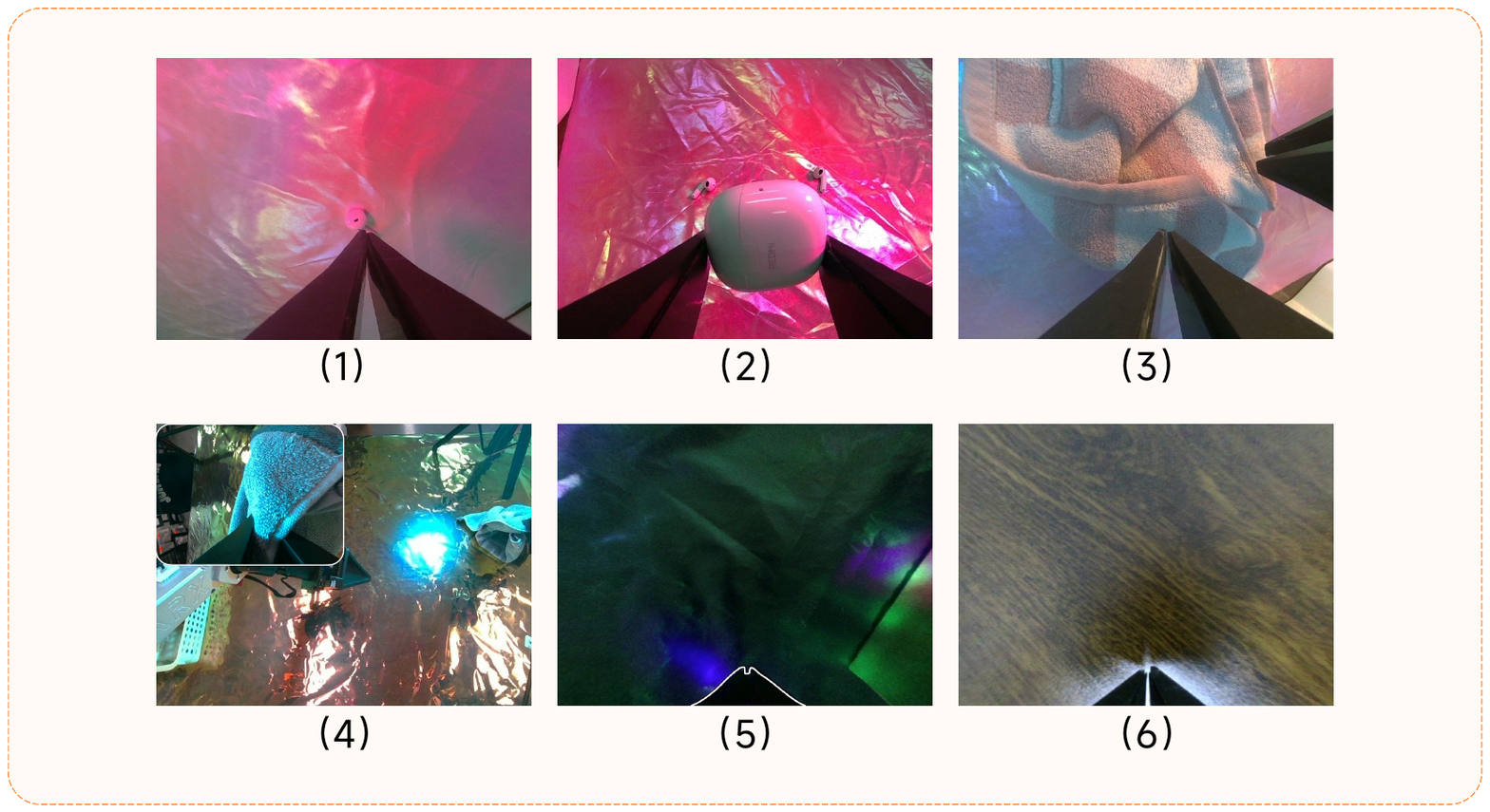}
    \caption{Typical failure cases across tasks: (1)--(2): \textit{Store Earphones}, (3)--(4): \textit{Fold Towel}, (5)--(6): \textit{Pack Box}.}
    \label{fig:failure_analysis}
\end{figure}

Figure~\ref{fig:failure_analysis} further shows the typical failure modes that arise in our evaluation:
(1)~failing to grasp the earbud under saturated lighting;
(2)~misidentifying the orientation of the earphone case;
(3)~failing to locate the towel edges against a visually similar tablecloth;
(4)~unfolding an already-folded towel due to an incomplete grasp of the cloth edge;
(5)(6)~attempting to grasp a projected light spot instead of the target object.
Under heavily corrupted visual observations, both policies occasionally exhibit these unexpected behaviors; however, \texttt{\Ours}-Aug is more likely to act correctly despite the perturbations or self-correct after a brief deviation.
The Original policy, by contrast, tends to become permanently stuck on visual artifacts or commit unrecoverable errors at a higher rate.

\begin{figure}[htbp]
\centering
\includegraphics[width=\columnwidth]{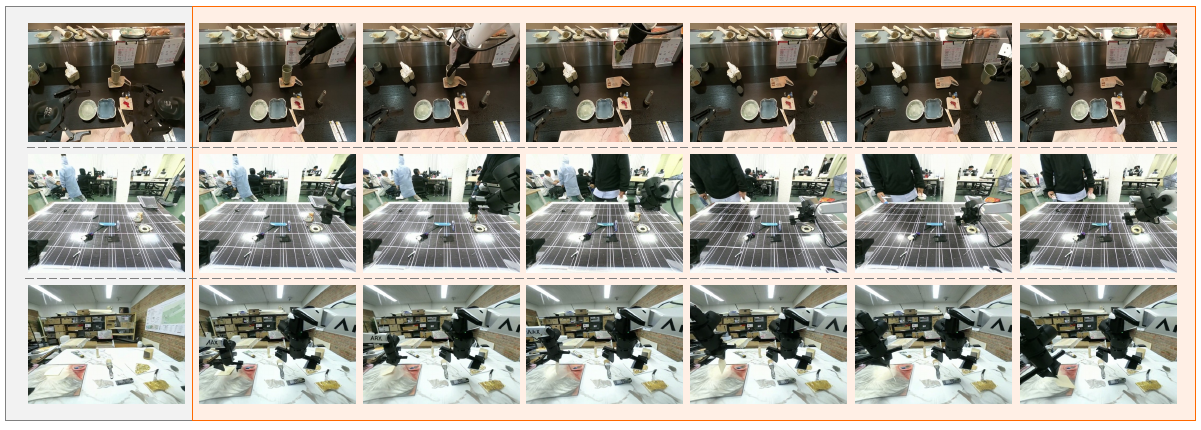}
\caption{Qualitative video generation results of \texttt{\Ours}.}
\label{fig:video_gen_vis_select}
\end{figure}


\subsection{Video Generation}
\begin{figure}[htbp]
\centering
\includegraphics[width=\columnwidth]{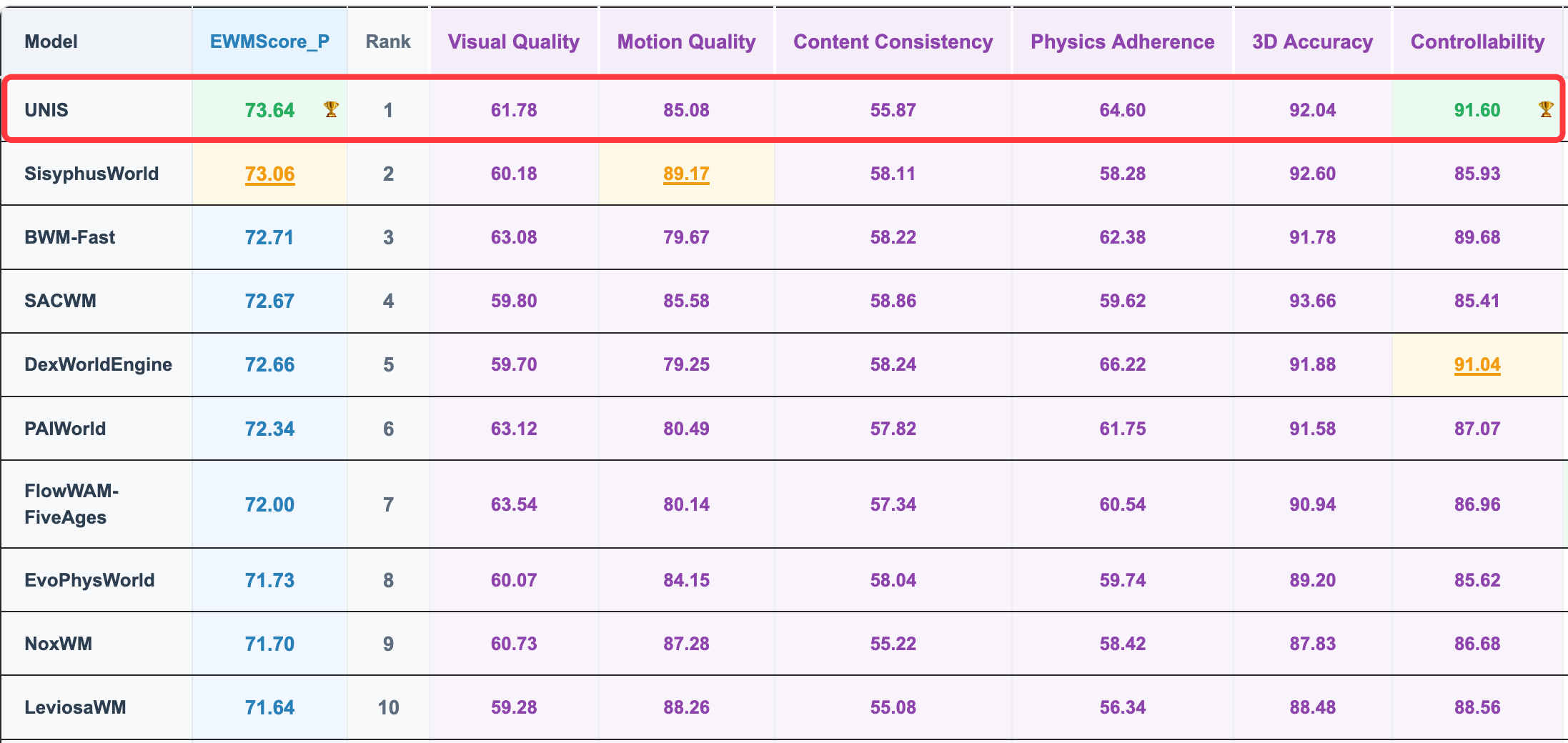}
\caption{WorldArena leaderboard (July 13, 2026). UNIS is the anonymous evaluation codename for \texttt{\Ours}.}
\label{fig:WorldArena_leaderboard}
\vspace{-10pt}
\end{figure}

\begin{figure*}[!t]
    \centering
    \includegraphics[width=\textwidth]{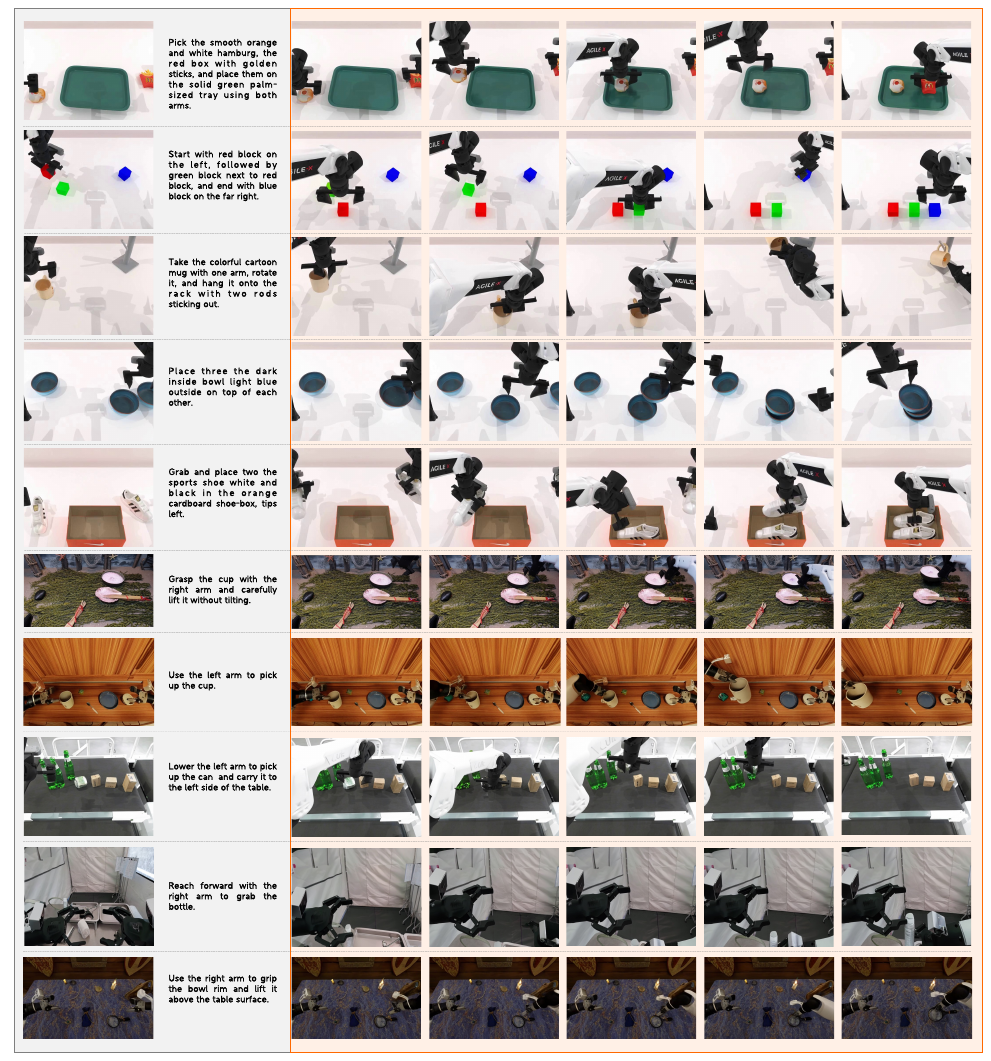}
    \caption{Qualitative video generation results of \Ours. Rows 1--5: WorldArena benchmark samples conditioned on language instructions and initial observations. Rows 6--10: videos generated from \Ours~produced scene images. Each row shows the initial frame and language instructions followed by predicted future frames.}
    \label{fig:video-gen}
\end{figure*}

\paragraph{WorldArena} A key capability of an Embodied World Model is serving as a data engine that generates physically plausible and visually high-quality videos from a single observation and a corresponding instruction. To evaluate this video generation capability for embodiment tasks, we benchmark \texttt{\Ours} on WorldArena~\cite{shang2026worldarena}. Given an initial frame along with a language instruction and robot action, WorldArena assesses the generated video across 15 metrics spanning six dimensions: visual quality, motion quality, content consistency, physics adherence, 3D accuracy, and controllability. We formulate this task as processing interleaved vision-language inputs and producing multi-frame interleaved outputs with empty language tokens. The initial frame paired with the language instruction serves as input, along with a sequence of robot mask images rendered from the robot action, camera parameters, and URDF as action guidance. Quantitative results are reported in Table~\ref{tab:worldarena} and Figure~\ref{fig:WorldArena_leaderboard}, qualitative visualizations are shown in Figure~\ref{fig:video-gen} (rows 1--5). \texttt{\Ours} achieves the best overall EWMScore (73.64), ranking first among over 100 submitted models. In terms of controllability, \texttt{\Ours} attains the highest Instruction Following (93.86) and ranks competitively on Semantic Alignment, demonstrating its strong ability to translate language instructions into faithful visual outcomes. For physics adherence, \texttt{\Ours} leads on Interaction Quality (87.30), indicating that it captures physically plausible object interactions. \texttt{\Ours} also achieves the best Perspectivity (98.84) and Motion Smoothness (95.51), reflecting accurate 3D viewpoint geometry and temporally stable motion. Overall, \texttt{\Ours} strikes the best balance across all evaluation dimensions, confirming its effectiveness as a data engine for embodiment tasks.

\paragraph{Embodiment Video Generation} 
Beyond the simulated WorldArena benchmark, we further leverage the scene generation capability of \texttt{\Ours} to produce novel initial frames and subsequently generate videos conditioned on language instructions. Representative examples are shown in Figure~\ref{fig:video_gen_vis_select}. In the first row, the robot arm correctly interacts with the cup—grasping it, lifting it, and placing it stably on the table, demonstrating accurate contact modeling. In the second row, beyond the correct tabletop manipulation, the model also renders dynamic background changes by simulating a person walking toward the table to observe the robot's operation, indicating awareness of scene-level dynamics beyond the immediate task. In the third row, the model faithfully simulates the physics of a deformable object (a sheet of paper): the paper bends realistically when picked up and set down, and exhibits a slight lag during lateral movement due to its low mass, reflecting an understanding of material-dependent physical properties. More visualizations are provided in Figure~\ref{fig:video-gen}.


\begin{table*}[htbp]
\vspace{-30pt}
\caption{WorldArena dimension-level scores and overall EWMScore. The best score in each column is \textbf{bolded}.}
\centering
\footnotesize
\setlength{\tabcolsep}{4pt}
\renewcommand{\arraystretch}{1.05}
\begin{tabular}{ll ccccc}
\toprule
\multicolumn{2}{l}{} & \texttt{\Ours} & SisyphusWorld & BWM-Fast & SACWM & DexWorldEngine \\
\midrule
\multirow{3}{*}{\parbox{1.8cm}{\centering Visual\\Quality}}
& Image Quality            & \textbf{53.94} & 45.57 & 51.22 & 45.83 & 51.44 \\
& Aesthetic Quality        & 40.79 & 38.38 & 40.15 & 37.30 & \textbf{40.96} \\
& JEPA Similarity          & 90.60 & 96.58 & \textbf{97.87} & 96.26 & 86.71 \\
\midrule
\multirow{3}{*}{\parbox{1.8cm}{\centering Motion\\Quality}}
& Dynamic Degree           & 73.70 & \textbf{76.27} & 69.58 & 70.35 & 69.16 \\
& Flow Score               & 86.02 & \textbf{99.82} & 75.11 & 95.79 & 74.31 \\
& Motion Smoothness        & \textbf{95.51} & 91.42 & 94.33 & 90.60 & 94.27 \\
\midrule
\multirow{3}{*}{\parbox{1.8cm}{\centering Content\\Consistency}}
& Subject Consistency      & 79.05 & 82.11 & 81.42 & \textbf{82.17} & 81.46 \\
& Background Consistency   & 86.44 & 86.52 & \textbf{90.17} & 87.90 & 88.85 \\
& Photometric Consistency  & 2.13 & 5.69 & 3.08 & \textbf{6.50} & 4.42 \\
\midrule
\multirow{2}{*}{\parbox{1.8cm}{\centering Physics\\Adherence}}
& Interaction Quality      & \textbf{87.30} & 71.98 & 79.88 & 72.28 & 81.04 \\
& Trajectory Accuracy      & 41.89 & 44.58 & 44.89 & 46.95 & \textbf{51.41} \\
\midrule
\multirow{2}{*}{\parbox{1.8cm}{\centering 3D\\Accuracy}}
& Depth Accuracy           & 85.25 & 94.85 & 86.41 & \textbf{94.99} & 86.19 \\
& Perspectivity            & \textbf{98.84} & 90.34 & 97.14 & 92.34 & 97.58 \\
\midrule
\multirow{2}{*}{\parbox{1.8cm}{\centering Controll-\\ability}}
& Instruction Following    & \textbf{93.86} & 82.68 & 90.22 & 81.66 & 92.62 \\
& Semantic Alignment       & 89.35 & 89.18 & 89.15 & 89.16 & \textbf{89.45} \\
\midrule
\multicolumn{2}{l}{\textbf{EWMScore} ($\uparrow$)}
                          & \textbf{73.64} & 73.06 & 72.71 & 72.67 & 72.66 \\
\bottomrule
\end{tabular}
\vspace{-5pt}
\label{tab:worldarena}
\end{table*}

\subsection{Text2Image \& Image Edit Benchmarks}

We evaluate the autoregressive \texttt{\Ours{}} image model on general text-to-image generation and image editing benchmarks. The goal of this evaluation is not to claim that \texttt{\Ours{}} is a dedicated open-domain image generator, but to quantify how much general image generation and editing ability is retained after adapting the model toward embodied scene synthesis. We use GenEval \cite{ghosh2023geneval} for compositional T2I evaluation and ImgEdit \cite{ye2026imgedit} Basic for single-turn image editing.

\begin{table*}[htbp]
\centering
\scriptsize
\setlength{\tabcolsep}{3pt}
\caption{Text-to-image and image-editing benchmark comparison with category-level breakdowns. GenEval scores are reported on a 0--1 scale, and ImgEdit scores use a 1--5 judge scale.}
\label{tab:t2i-imgedit-results}
\resizebox{\textwidth}{!}{%
\begin{tabular}{@{}lccccccc@{}}
\toprule
Model & Overall & Single object & Two object & Counting & Colors & Position & Attribute binding \\
\midrule
Qwen-Image & 0.87 & 0.99 & 0.92 & 0.89 & 0.88 & 0.76 & 0.77 \\
\Ours{}-AR & 0.74 & 0.99 & 0.92 & 0.49 & 0.88 & 0.60 & 0.53 \\
\bottomrule
\end{tabular}
\vspace{-20pt}
}

\vspace{0.4em}

\resizebox{\textwidth}{!}{%
\begin{tabular}{@{}lcccccccccc@{}}
\toprule
Model & Overall & Add & Adjust & Extract & Replace & Remove & Background & Style & Hybrid & Action \\
\midrule
Qwen-Image & 4.27 & 4.38 & 4.16 & 3.43 & 4.66 & 4.14 & 4.38 & 4.81 & 3.82 & 4.69 \\
\Ours{}-AR & 4.08 & 4.21 & 4.15 & 3.28 & 4.35 & 4.21 & 4.03 & 4.72 & 3.27 & 4.53 \\
\bottomrule
\end{tabular}
}
\vspace{-10pt}
\end{table*}

As shown in Table~\ref{tab:t2i-imgedit-results}, \texttt{\Ours{}} preserves strong general T2I and image-editing capabilities after embodied post-training. It performs competitively on most appearance-oriented generation and local editing tasks. Performance drops are largely confined to highly complex scenarios that require multi-constraint composition, precise spatial grounding, or symbolic binding.

Table~\ref{tab:t2i-imgedit-results} demonstrates that \texttt{\Ours{}} retains robust general T2I and image-editing capabilities after embodied post-training. While it slightly lags behind Qwen-Image in highly complex cases—such as multi-constraint composition or precise spatial alignment—\texttt{\Ours{}} maintains highly comparable performance on common image generation and local editing instructions without catastrophic forgetting.

\subsection{Model Distillation}
To reduce the computational cost of Xiaomi-Robotics-U0 while retaining its multimodal generation capabilities, we distill the 34B autoregressive model into a compact 4B student. The student follows the same decoder-only autoregressive formulation, tokenizer, multimodal vocabulary, and visual tokenization scheme as the teacher, allowing it to support text-to-image generation, image editing, and interleaved embodied-video generation within a unified architecture. The 4B model contains 36 Transformer layers with a hidden dimension of 2,560, an FFN dimension of 9,728, and 32 attention heads organized into 8 groups. It supports sequences of up to 32,768 tokens, which is sufficient for long interleaved sequences containing language instructions, control tokens, and multiple image frames.

We employ token-level knowledge distillation by matching the temperature-scaled output distributions of the frozen 34B Xiaomi-Robotics-U0 teacher and the student \cite{hinton2015distilling}. During distillation, the frozen teacher and the student receive the same ground-truth multimodal token sequence and predict the next-token distributions under identical contexts. Gradients are propagated only through the student. The student is optimized using both ground-truth next-token supervision and the teacher’s soft output distribution. Specifically, the training objective is

\begin{equation}
\mathcal{L}_{\mathrm{distill}}
=
\mathcal{L}_{\mathrm{img}}
+
\mathcal{L}_{\mathrm{text}}
+
\lambda_{\mathrm{KD}} T^{2}
D_{\mathrm{KL}}
\left(
p_{\mathrm{teacher}}^{(T)}
\parallel
p_{\mathrm{student}}^{(T)}
\right),
\label{eq:distillation-objective}
\end{equation}

where $\mathcal{L}_{\mathrm{img}}$ and $\mathcal{L}_{\mathrm{text}}$ denote token-mean cross-entropy losses computed separately over supervised visual-token positions and supervised language, control, and boundary-token positions, respectively. $\lambda_{\mathrm{KD}}=0.3$ is the weight of the knowledge distillation loss, and $D_{\mathrm{KL}}(\cdot\parallel\cdot)$ denotes the Kullback--Leibler divergence. Specifically, $p_{\mathrm{teacher}}^{(T)}$ and $p_{\mathrm{student}}^{(T)}$ are the temperature-scaled output distributions of the teacher and student, respectively, defined as $p^{(T)}=\operatorname{softmax}(z/T)$, where $z$ denotes the corresponding logits and $T=2$ is the temperature. The supervision mask excludes the conditioning prefix and covers the complete target span, including visual tokens as well as language, control, and special boundary tokens. Ground-truth cross-entropy is computed separately for visual-token positions and for language/control-token positions, with each component normalized by the number of tokens in its own group. This prevents the substantially larger number of visual tokens from dominating the supervision of sparse language, control, and boundary tokens. The KL divergence is computed at supervised visual-token positions over the complete multimodal output vocabulary. Therefore, the student learns the teacher's probability distribution over all possible multimodal tokens.

During training, we employ a global batch size of 1,024 and a learning rate of \(5\times10^{-5}\), with a 1\% warm-up phase followed by cosine decay. The attention and hidden-state dropout rates are both set to 0.1. Distillation is performed for the first 2,000 optimization steps, after which the frozen teacher is released and training proceeds with the ground-truth cross-entropy objective. The subsequent training stage runs for 5,000 steps for the single-step generation model and 3,000 steps for the interleaved video-generation model. This schedule provides teacher guidance during the initial adaptation stage, followed by direct optimization on the target training distribution.

\begin{figure*}[htbp]
    \centering
    \includegraphics[width=\textwidth]{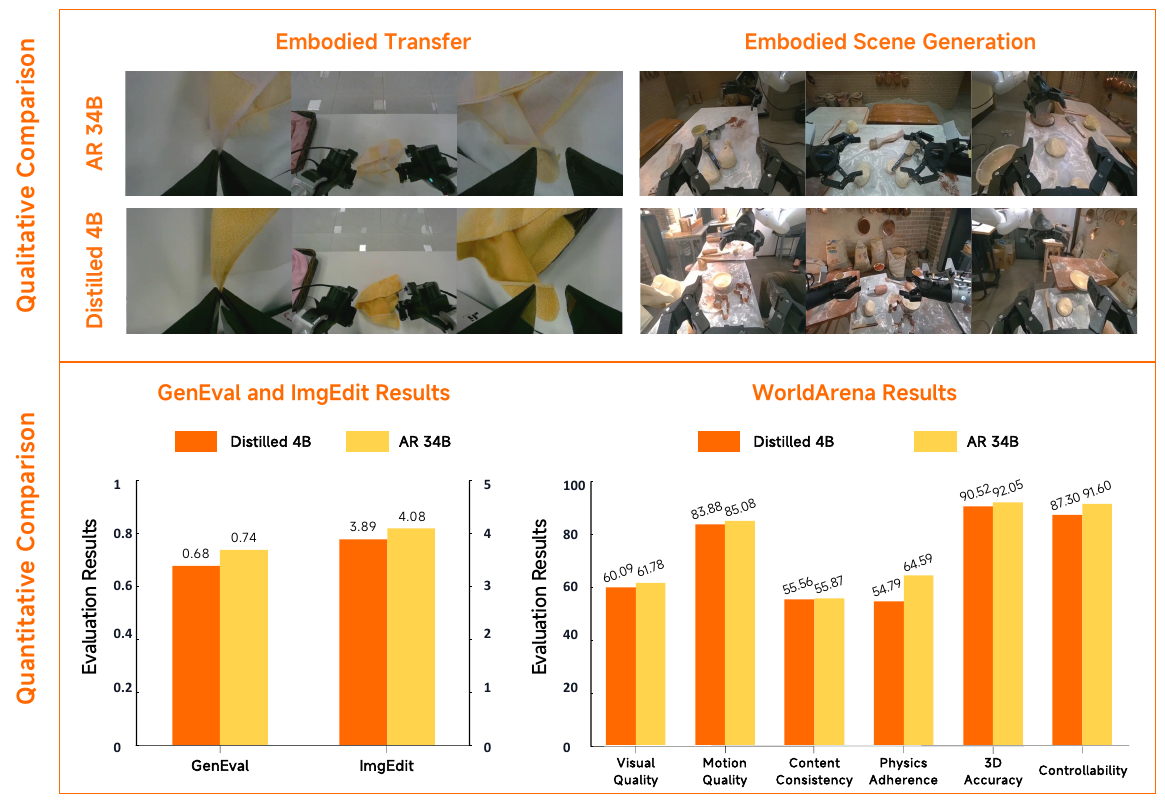}
    \caption{Qualitative and quantitative comparison between the distilled 4B student and the autoregressive 34B Xiaomi-Robotics-U0 teacher.}
    \label{fig:distilled_4b}
\end{figure*}

The qualitative comparisons in the upper part of Figure~\ref{fig:distilled_4b} show that the 4B student retains the teacher’s core embodied-generation behavior. In the embodied-transfer examples, it preserves the target scene structure and robot--object configuration across views. In embodied scene generation, it produces coherent robot workspaces while maintaining the semantic content and overall spatial organization of the conditioning input.

Despite an approximately \(8.5\times\) reduction in parameter count, the distilled model achieves competitive performance on general image generation and editing benchmarks, as shown in the lower-left panel of Figure~\ref{fig:distilled_4b}. On GenEval~\cite{ghosh2023geneval}, the 4B model obtains 0.68 compared with 0.74 for the 34B teacher. On ImgEdit~\cite{ye2026imgedit}, it achieves 3.89 versus 4.08. These results indicate that the distilled model maintains strong single-step image generation and editing capabilities.

The lower-right panel of Figure~\ref{fig:distilled_4b} summarizes the model's performance on WorldArena~\cite{shang2026worldarena}. For each of the six evaluation dimensions, we report the average score of its constituent metrics. Overall, the distilled 4B model achieves performance close to the 34B teacher in visual quality, motion quality, content consistency, and 3D accuracy. The main performance gaps occur in physics adherence and controllability, suggesting that interaction reasoning and precise instruction-conditioned behavior benefit more from the larger model capacity. The results show that the distilled model provides a substantially more compact alternative while maintaining strong performance across the major embodied-generation dimensions.

\section{Related Work}
\label{related_work}

\subsection{Foundation Generative Models and Embodied World Models}

Recent years have witnessed rapid advances in foundation generative models, evolving from image synthesis to unified multimodal world modeling. Large-scale image generation and multimodal generation models, including Stable Diffusion~\cite{blattmann2023stable}, FLUX~\cite{labs2025flux1kontextflowmatching}, GPT-Image-2.0~\cite{openai_gptimage_2025}, Show-o~\cite{xie2025show}, Janus~\cite{wu2025janus}, Janus-Pro~\cite{chen2025janus}, Chameleon~\cite{team2024chameleon}, BAGEL~\cite{deng2025emerging}, and EMU3.5~\cite{cui2025emu3}, have demonstrated remarkable capabilities in semantic understanding, instruction following, compositional reasoning, and controllable visual generation. More recently, foundation video and world models, such as Genie~\cite{bruce2024genie}, Genie2~\cite{parkerholder2024genie2}, Sora~\cite{openai2024sora}, MovieGen~\cite{polyak2024movie}, Wan~\cite{wan2025wan}, and Cosmos~\cite{agarwal2026cosmos}, further extend generation from static images to long-horizon videos and interactive world simulation, enabling coherent prediction of future observations under a unified autoregressive or diffusion framework. Despite their impressive generalization ability, these models are designed for generic visual content and do not explicitly model the geometric constraints, robot embodiments, calibrated camera systems, or physical interaction dynamics required by embodied manipulation.

To bridge this gap, recent embodied world models introduce robot-centric world modeling by adapting foundation models to manipulation trajectories and robot videos. Representative approaches include Dreamer Series~\cite{hafner2019dream}, World Action Models (WAMs)~\cite{yuan2026fast, ye2026gigaworld, li2026causal, team2026motubrain}, DreamZero~\cite{ye2026world}, and Qwen-RobotWorld~\cite{zhang2026qwen}, which learn predictive models of future observations for planning, simulation, or policy learning. While these methods significantly advance embodied prediction, they are typically optimized using robot-only datasets whose scale and diversity remain substantially smaller than those used for foundation generative models. Consequently, continual adaptation often sacrifices part of the semantic richness, controllability, and visual diversity inherited from foundation pre-training. In contrast, our work treats embodied generation as a natural extension of foundation image and video generation. Rather than specializing a foundation model solely on robot trajectories, we jointly optimize general-domain generation and embodied synthesis within a unified autoregressive framework, preserving the original generation capability while acquiring robot-centric geometric reasoning and interaction modeling.

\subsection{Controllable and Multi-view Embodied Generation}

Recent advances in controllable image generation and image editing have substantially improved semantic manipulation through instruction-guided synthesis and reference-conditioned editing. Representative approaches, including InstructPix2Pix~\cite{brooks2023instructpix2pix}, AnyDoor~\cite{chen2024anydoor}, OmniGen~\cite{xiao2025omnigen}, GPT-Image-2.0~\cite{openai_gptimage_2025}, and FLUX Kontext~\cite{labs2025flux1kontextflowmatching}, enable high-quality editing with strong instruction following and semantic consistency. However, these methods are primarily designed for natural images and generally assume single-view observations, making them difficult to directly apply to embodied manipulation, where multiple calibrated cameras, robot embodiments, and interaction states must remain geometrically consistent.

Meanwhile, multi-view generation has attracted increasing attention in image synthesis and 3D content creation. Representative methods, including MVDream~\cite{shi2024mvdream}, SyncDreamer~\cite{liu2024syncdreamer}, Wonder3D~\cite{long2024wonder3d}, Era3D~\cite{li2024era3d}, Zero123++~\cite{shi2023zero123++}, and related multi-view diffusion models~\cite{gao2024cat3d,tang2024mvdiffusion++}, explicitly enforce cross-view consistency to facilitate high-quality 3D reconstruction and asset generation. Nevertheless, these methods mainly focus on object-centric generation and do not explicitly consider robot embodiments, calibrated manipulation cameras, or interaction-aware scene synthesis.

Our work bridges these two research directions by unifying controllable image editing, multi-view embodied scene generation, and embodied transfer within a single autoregressive framework. Furthermore, we introduce a structured control formulation that disentangles workspace layout, background appearance, foreground irrelevant objects, target objects, and lighting into independent control dimensions, enabling significantly richer embodied scene variations while preserving robot geometry, embodiment constraints, and multi-view consistency.

\subsection{Embodied Video Generation and Synthetic Data Engines}

Video generation and embodied world simulation have recently become increasingly important for scalable robot learning. Foundation video models such as MovieGen~\cite{polyak2024movie}, Wan~\cite{wan2025wan}, Cosmos~\cite{agarwal2026cosmos}, Genie~\cite{bruce2024genie}, and Qwen-RobotWorld~\cite{zhang2026qwen} demonstrate impressive capabilities in long-horizon video synthesis and future observation prediction. Meanwhile, large-scale robot datasets and simulation platforms, including Open X-Embodiment~\cite{o2024open}, RoboCasa~\cite{nasiriany2024robocasa}, ManiSkill~\cite{mu2021maniskill}, BridgeData V2~\cite{walke2023bridgedata}, and DROID~\cite{khazatsky2024droid}, have substantially accelerated the development of data-driven robot learning. More recently, embodied video datasets and synthetic data pipelines, such as RoboMIND~\cite{wu2024robomind}, RoboCoin~\cite{wu2025robocoin}, GenieSim~\cite{yin2026geniesim30}, InternA1~\cite{tian2026interndata}, and AgiBotWorld~\cite{bu2025agibot}, further explore large-scale embodied video generation and robot data augmentation.

Despite these advances, existing approaches typically synthesize isolated images, simulation assets, or short robot videos independently, without jointly modeling controllable scene generation, embodied transfer, and long-horizon trajectory rollout. In contrast, our work unifies embodied scene generation, embodied transfer, interleaved subtask-subgoal prediction, and embodied video generation within a single autoregressive model. By jointly learning embodied videos at multiple temporal resolutions, our model captures both long-horizon task progression and fine-grained interaction dynamics, transforming foundation world models into scalable embodied data engines for downstream policy learning and future agentic embodied systems.
\section{Conclusions}
\label{conclusions}
In this report, we presented \texttt{\Ours}, a unified embodied synthesis model built upon a world foundation model through lightweight continual training on general-domain and embodied datasets. \texttt{\Ours} unifies text-to-image generation, image editing, multi-view embodied scene generation, embodied transfer, and embodied video generation within a single autoregressive framework, successfully transferring the rich visual knowledge of foundation models into embodied intelligence. To the best of our knowledge, \texttt{\Ours} is the first unified embodied synthesis model supporting both high-quality multi-view embodied generation and sequential embodied video generation, as well as the first to introduce a structured control formulation for scalable embodied data generation. Extensive experiments demonstrate state-of-the-art performance across both single-step and sequential embodied generation tasks, outperforming GPT-Image-2.0 in human evaluation, achieving state-of-the-art results on the World Arena benchmark, and substantially improving the robustness of downstream robot policies through generated data. We hope this work provides a practical step toward scalable agentic data engines for embodied intelligence.

Despite the promising results, \texttt{\Ours} has several limitations. First, embodied transfer still relies on depth estimation as an intermediate representation, which may introduce occasional artifacts and limit fine-grained control over object textures and local appearances. Future work will investigate embodied multi-view image editing directly on raw observations. Second, embodied scene generation and video generation are performed separately, making long-horizon rollout susceptible to accumulated errors; jointly generating scenes and subsequent videos is a promising direction. Finally, the current 32K context window limits long-duration video modeling, and extending the model to longer contexts will enable minute-level embodied interaction generation.

\clearpage

\section*{Contributions}
\setlength{\parskip}{5pt} 
\setlength{\itemsep}{0pt} 
\label{contributions}
$^*$Core contributors, alphabetically listed by last name ~~~$^{**}$External contributor \\

\begin{multicols}{-2}

\textbf{Project Lead}
\begin{itemize}
    \item Xinghang Li
\end{itemize}

\textbf{Data}: 
\begin{itemize}
    \item Jun Guo$^*$
    \item Qiwei Li$^*$
    \item Hang Lai$^*$
    \item Long Qian$^*$
    \item Yueze Wang$^*$
    \item Nan Sun$^*$
    \item Hongyu Yan$^*$
    \item Jingen Qu
    \item Jiaxi Song
    \item Jiahang Cao
    \item Hanye Zhao
    \item Heyun Wang
    \item Yunhong Wang
    \item Jiaxi Song
    \item Caoyu Xia
    \item Jack Zhao
\end{itemize}

\textbf{Training}: 
\begin{itemize}
    \item Jun Guo$^*$
    \item Qiwei Li$^*$
    \item Long Qian$^*$
    \item Yueze Wang$^*$
    \item Jingen Qu
    \item Jiaxi Song
    \item Nan Sun
    \item Hongyu Yan
    \item Yingbo Tang
\end{itemize}

\textbf{Deployment \& Experiments}: 
\begin{itemize}
    \item Jiahang Cao$^*$
    \item Jun Guo$^*$
    \item Hang Lai$^*$
    \item Qiwei Li$^*$
    \item Long Qian$^*$
    \item Jiaxi Song$^*$
    \item Hongyu Yan$^*$
    \item Hanye Zhao$^*$
    \item Futeng Liu
    \item Wanli Peng
\end{itemize}

\textbf{Inference Speedup}: 
\begin{itemize}
    \item Xi Chen$^*$
    \item Jingen Qu$^*$
\end{itemize}

\textbf{Infrastructure}: 
\begin{itemize}
    \item Xinghang Li$^*$
    \item Jingen Qu
    \item Xi Chen
    \item Yingbo Tang
\end{itemize}

\textbf{Senior Lead}: 
\begin{itemize}
    \item Jason Li$^*$
    \item Diyun Xiang
    \item Hangjun Ye
    \item Heng Qu
    \item Huaping Liu$^{**}$
\end{itemize}

\end{multicols}

\section*{Acknowledgment}
\label{Acknowledgment}

We would like to express our sincere appreciation to the broader team for their tremendous support, including those not listed above: Li Jiang, Zhiyin Lai, Xiaoke Xilinjueluo, Qingyi Li, Qi Liu, Yayun Liu, Jun Xia, Feng Qiu, Donghao Wang, Yan Hou, Dong Wang, Quanyun Zhou, Liangliang He, Jiaxin Liu, Kang Zhou, Rui Cai, Shuoxue Bi, Yingchao Zhou, Kun Ma, Yiwei Zhou

\begin{figure*}[!t]
    \centering
    \includegraphics[width=\textwidth]{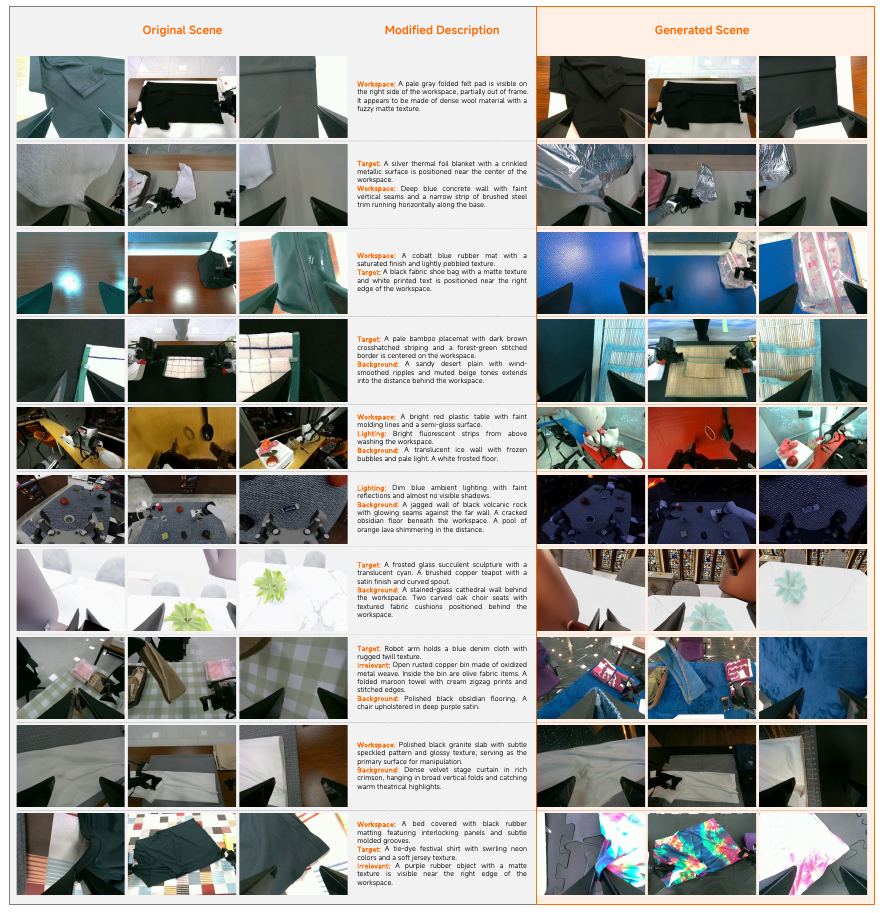}
    \caption{Qualitative results of embodied transfer. We extract depths and modify descriptions from the original scene to generate the new scene.}
    \label{fig:transfer-full}
\end{figure*}

\begin{figure*}[t]
    \centering
    \includegraphics[width=1\linewidth]{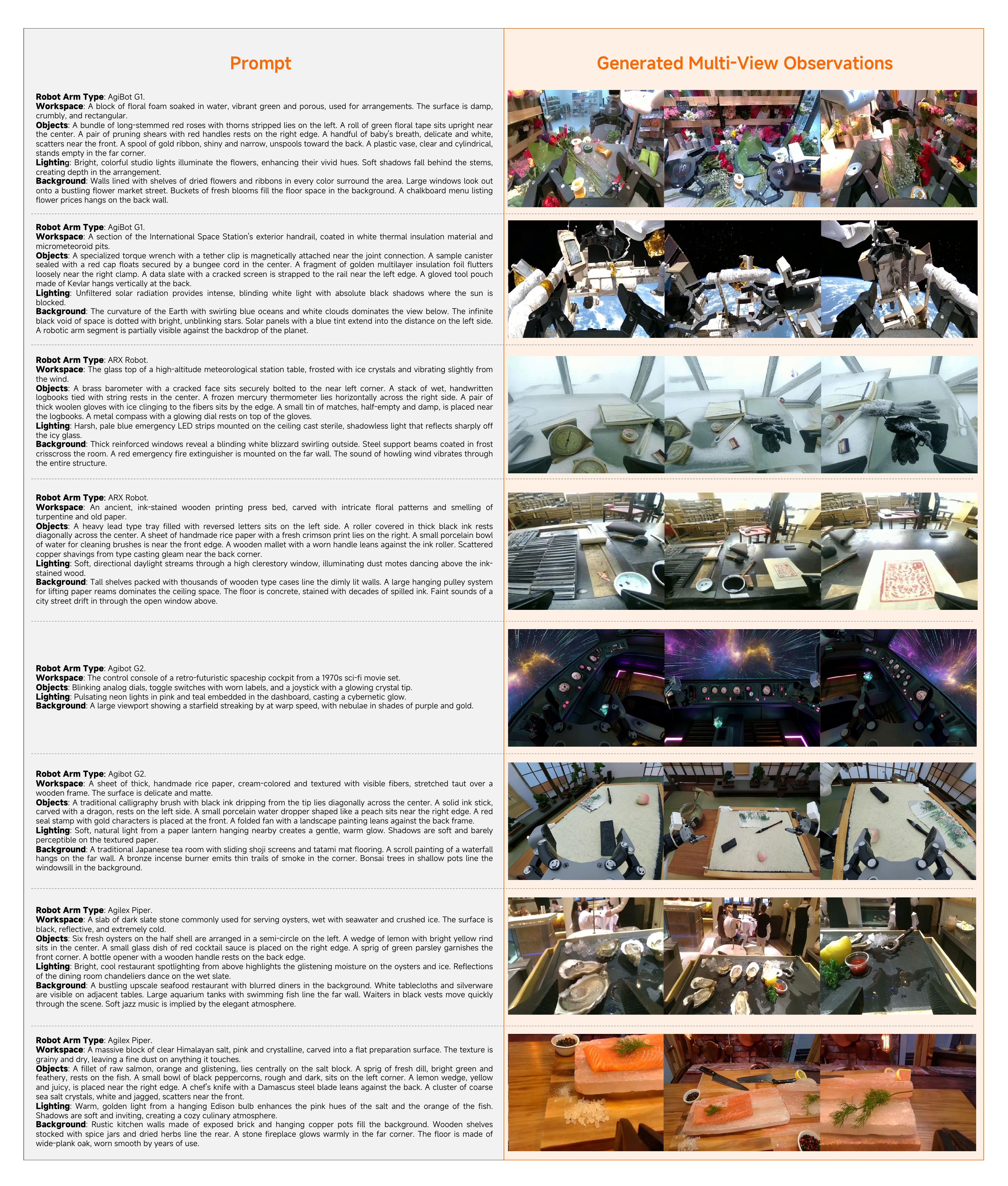}
    \caption{Embodied scene generation results.}
    \label{fig:scene_gen_examples}
\end{figure*}

\begin{figure*}[!t]
    \centering
    \includegraphics[width=\textwidth]{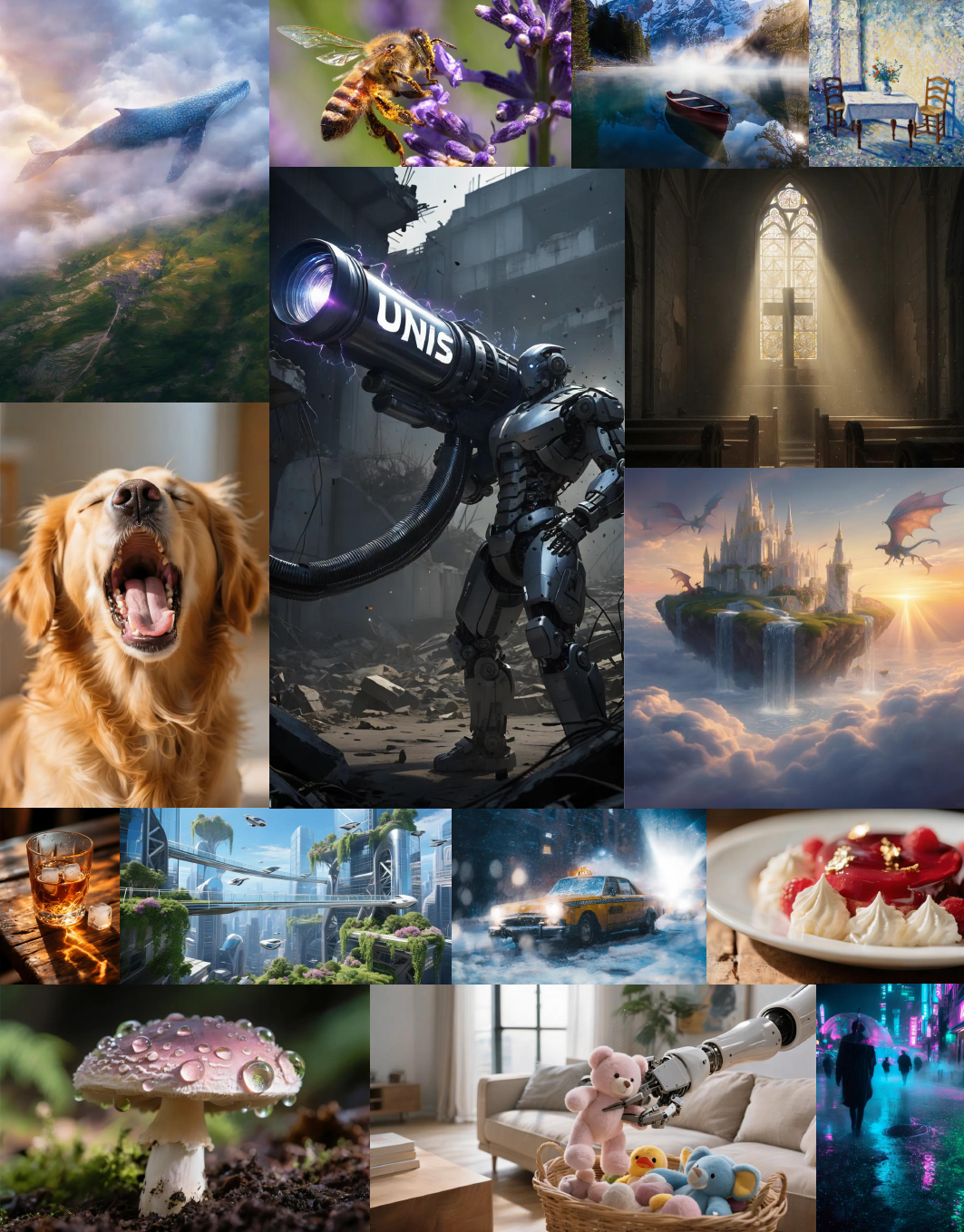}
    \caption{Text-to-image results of \Ours.}
    \label{fig:t2i_examples}
\end{figure*}

\begin{figure*}[!t]
    \centering
    \includegraphics[width=0.9\textwidth]{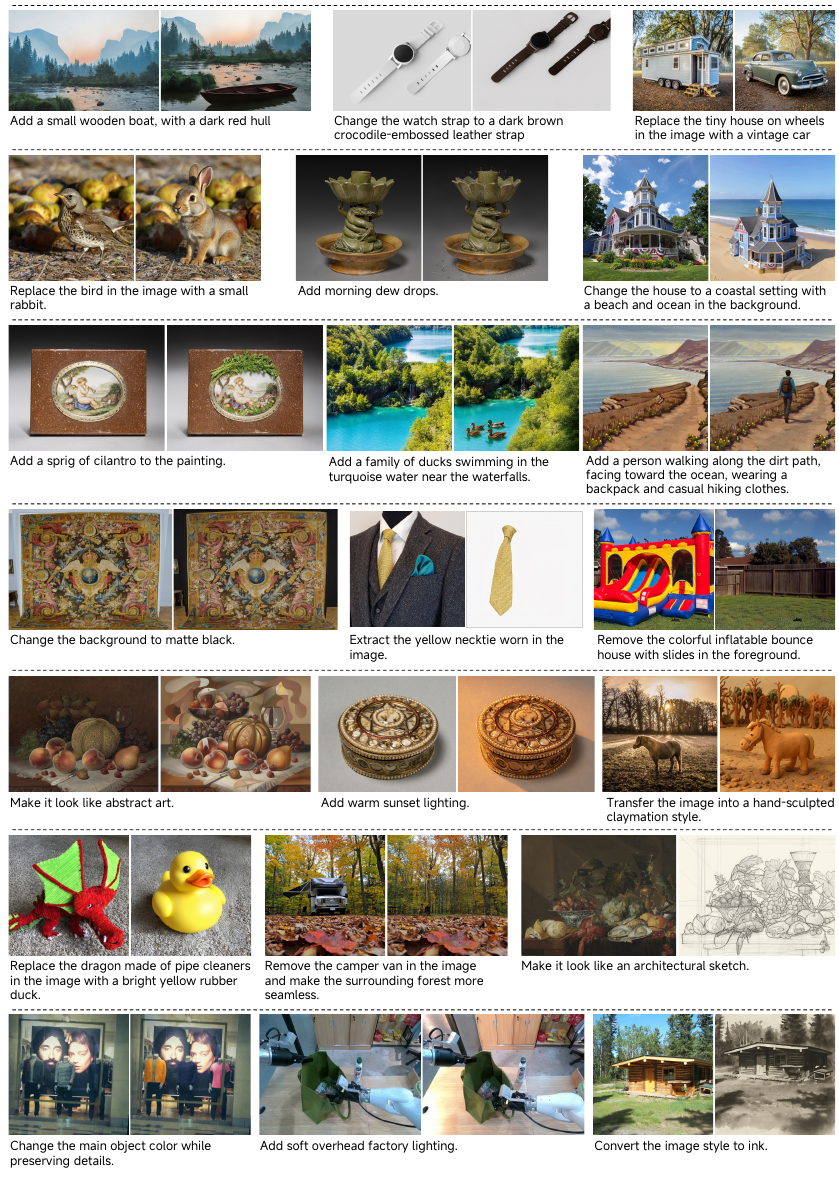}
    \caption{Image editing results of \Ours.}
    \label{fig:x2i_examples}
\end{figure*}

\clearpage
\bibliographystyle{plainnat}
\bibliography{main}



\end{document}